\documentclass{article}

    \PassOptionsToPackage{numbers, compress}{natbib}

\usepackage[preprint]{neurips_2024}

\usepackage[utf8]{inputenc} %
\usepackage[T1]{fontenc}    %
\usepackage[hidelinks,colorlinks=true,linkcolor=blue,citecolor=blue]{hyperref}        %
\usepackage{url}            %
\usepackage{booktabs}       %
\usepackage{amsfonts}       %
\usepackage{nicefrac}       %
\usepackage{microtype}      %
\usepackage{xcolor}         %
\usepackage{tcolorbox}
\usepackage{amsmath}
\usepackage{soul} 
\usepackage{multirow}
\usepackage{booktabs}
\usepackage{adjustbox}
\usepackage{hhline}
\usepackage{tabularx} 
\usepackage{subcaption}
\usepackage{enumitem}
\usepackage{wrapfig}
\usepackage{algorithm}
\usepackage{algpseudocode}
\usepackage{colortbl}
\usepackage{tcolorbox}
\usepackage{xparse}
\usepackage{stfloats}
\usepackage{highlight}
\usepackage{adjustbox}
\usepackage{graphicx}

\usepackage{xspace}

\newcommand{\homepage}{\raisebox{-1.5pt}{\includegraphics[height=1.05em]{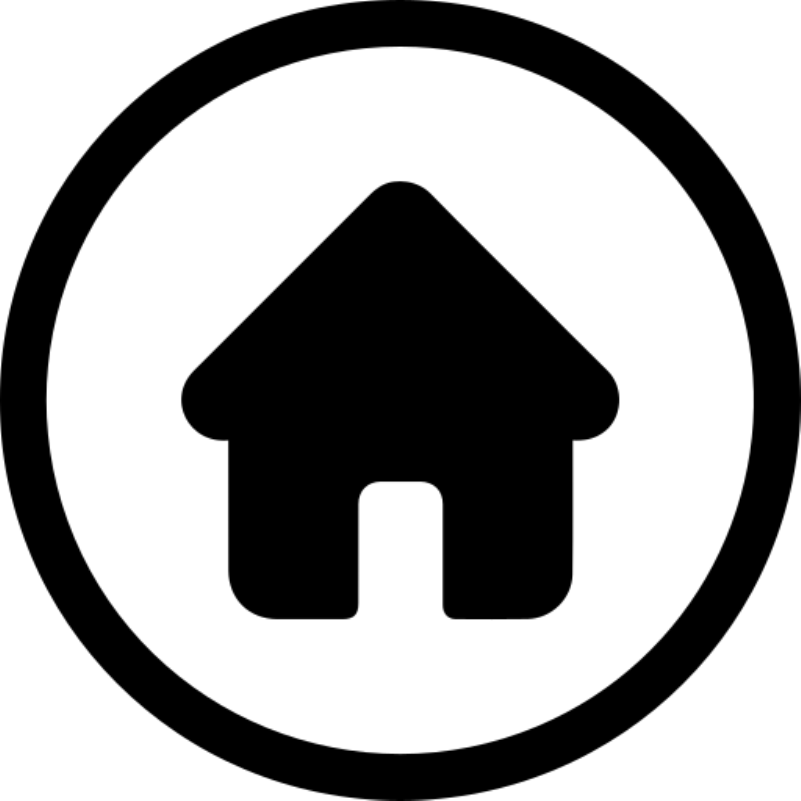}}\xspace}

\newcommand{\huggingface}{\raisebox{-1.5pt}{\includegraphics[height=1.05em]{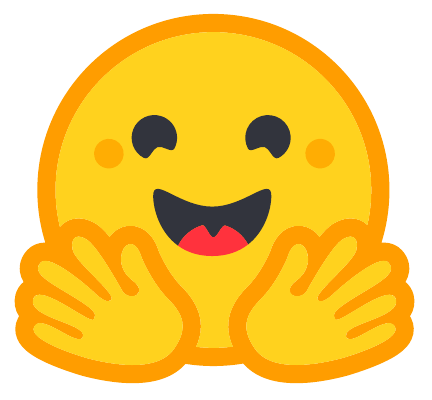}}\xspace}
\newcommand{\database}{\raisebox{-1.5pt}{\includegraphics[height=1.05em]{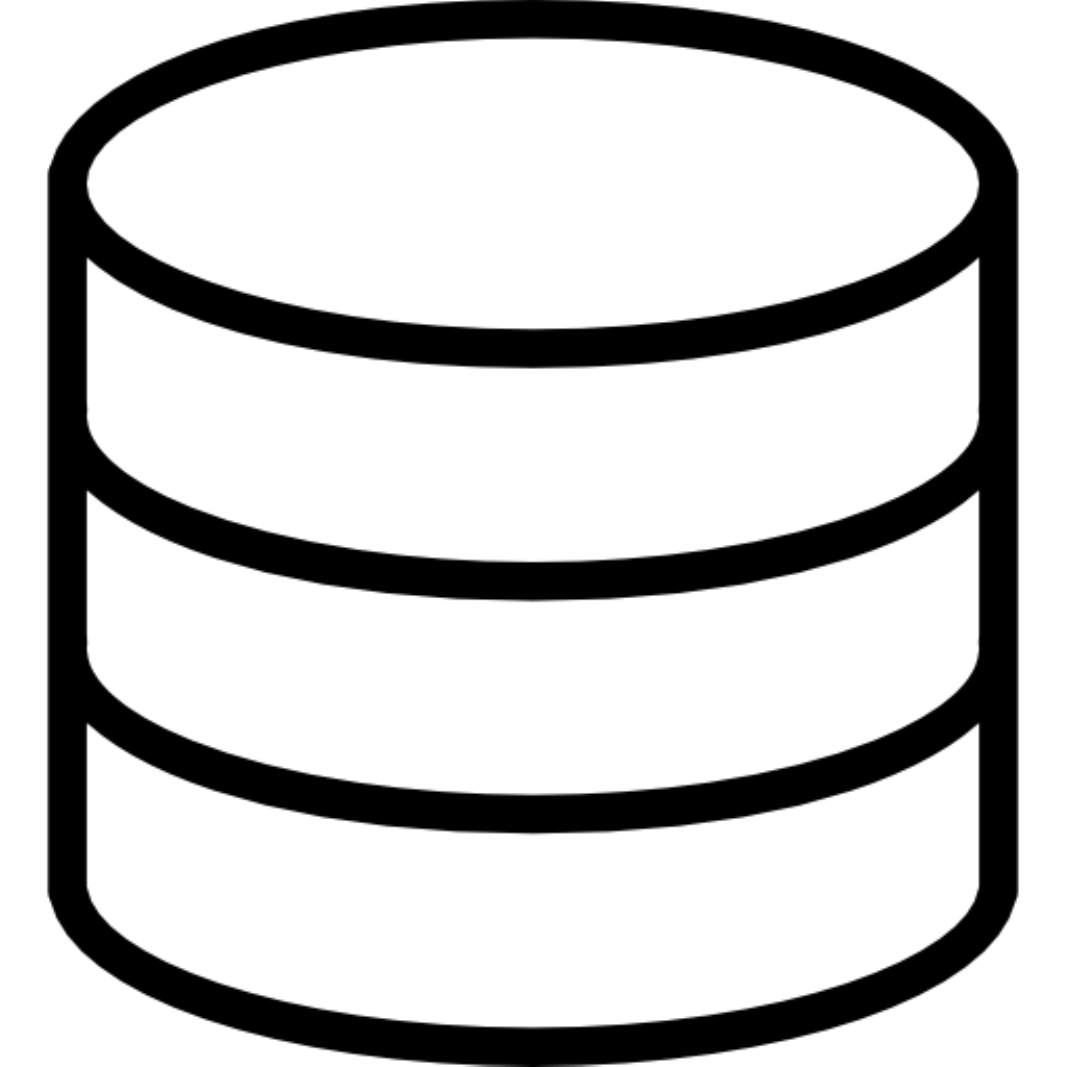}}\xspace}

\newcommand{\pipeline}{\texttt{APIGen-MT}}
\newcommand{\modelname}{\texttt{xLAM-2-70b-fc-r}}
\newcommand{\modelseries}{\texttt{xLAM-2-fc-r}}

\title{\pipeline: \underline{A}gentic \underline{PI}peline for \underline{M}ulti-\underline{T}urn Data \underline{Gen}eration via Simulated Agent-Human Interplay }

\author{
\bfseries{Akshara Prabhakar\thanks{Co-first Authors}  ~~Zuxin Liu\footnotemark[1] ~~Ming Zhu\thanks{Core Contributors} ~~Jianguo Zhang\footnotemark[2] ~~Tulika Awalgaonkar\footnotemark[2]}\\
\bfseries{Shiyu Wang ~~Zhiwei Liu ~~Haolin Chen ~~Thai Hoang ~~Juan Carlos Niebles}\\
\bfseries{Shelby Heinecke\thanks{Corresponding Authors}\footnotemark[3] ~~Weiran Yao\footnotemark[3] ~~Huan Wang\footnotemark[3] ~~Silvio Savarese\footnotemark[3] ~~Caiming Xiong\footnotemark[3]}\\\\
Salesforce AI Research\\
}

\begin{document}

\maketitle

\begin{abstract}
  Training effective AI agents for multi-turn interactions requires high-quality data that captures realistic human-agent dynamics, yet such data is scarce and expensive to collect manually. We introduce \pipeline, a two-phase framework that generates verifiable and diverse multi-turn agent data. In the first phase, our agentic pipeline produces detailed task blueprints with ground-truth actions, leveraging a committee of LLM reviewers and iterative feedback loops. These blueprints are then transformed into complete interaction trajectories through simulated human-agent interplay. We train a family of models---the \modelseries{} series with sizes ranging from 1B to 70B parameters.
Our models outperform frontier models such as GPT-4o and Claude 3.5 on $\tau$-bench and BFCL benchmarks, with the smaller models surpassing their larger counterparts, particularly in multi-turn settings, while maintaining superior consistency across multiple trials. 
Comprehensive experiments demonstrate that our verified blueprint-to-details approach yields high-quality training data, enabling the development of more reliable, efficient, and capable agents. We open-source 5K synthetic data trajectories and the trained \modelseries{} models to advance research in AI agents.

\begin{center}
\begin{tabular}{rll}
    \huggingface & \textbf{\small{Model}} & 
    \href{https://huggingface.co/collections/Salesforce/xlam-2-67ef5be12949d8dcdae354c4}{\texttt{https://huggingface.co/Salesforce/xLAM-2}}\\
    \database & \textbf{\small{Dataset}} & \href{https://huggingface.co/datasets/Salesforce/APIGen-MT-5k}{\texttt{https://huggingface.co/Salesforce/APIGen-MT-5k}}\\
    \homepage & \textbf{\small{Website}} & \href{https://apigen-mt.github.io/}{\texttt{https://apigen-mt.github.io}}
\end{tabular}
\end{center}

\end{abstract}

\section{Introduction}
\label{sec:introduction}

The growth of Large Language Model (LLM) agents has been accelerating at an unprecedented rate, driven by advancements in AI capabilities and increasing demand across various industries \cite{li2023camel,agashe2024agent_s,owl2025,antoniades2024swe,pan2024training,zhang2024diversity,li2024benchmarking,bahdanau2024tapeagents,liu2023tail}. Their role has evolved beyond simple conversational chatbots to AI agents capable of executing real-world tasks, such as managing financial transactions, scheduling appointments, and handling customer service requests. These applications demand not only linguistic fluency but also precise execution, reliability, and adherence to domain-specific policies.
Realistic enterprise use cases involve having an assistant (also referred to as \textit{agent} in this document) that is capable of fluently conversing with humans of different personalities, incrementally understanding their intent, extracting the background details needed, accurately invoke APIs, and operate over a complex business logic structure.

\begin{figure}[t]
\centering
\includegraphics[width=\linewidth]{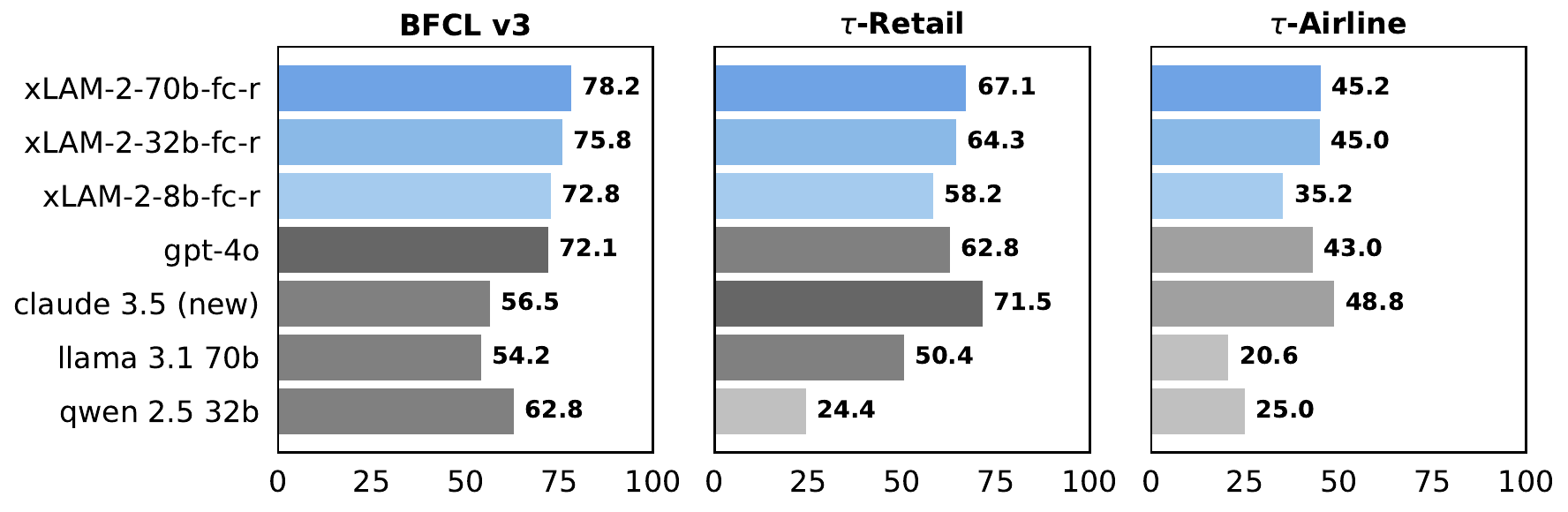}
\caption{\small{Comparative performance of larger \modelseries~ models (8B-70B, trained with \pipeline~ data) against state-of-the-art baselines on function-calling (BFCL v3 \cite{berkeley-function-calling-leaderboard}) and agentic ($\tau$-bench \cite{yao2024tau}) capabilities.}}

\label{fig:model_leaderboard}
\end{figure}

Despite their potential, building robust and reliable AI agents presents significant challenges \cite{yao2024tau}. Recent benchmarks reveal that even advanced LLMs struggle with multi-turn interactions, particularly when required to perform complex function calls, track long-term dependencies, or request missing information \cite{yin2025magnetmultiturntoolusedata,lu2024toolsandbox,yang2023language,berkeley-function-calling-leaderboard,huang2025crmarenaunderstandingcapacityllm}. Although framework design and prompt engineering have shown promise, the underlying model capabilities remain the primary bottleneck, largely due to two fundamental obstacles: (1) the scarcity of high-quality agent interaction data in public pretraining corpora, and (2) the prohibitive cost and time required to manually collect and label such data, especially for domain-specific applications requiring specialized knowledge.

Several approaches have attempted to address these challenges. APIGen \cite{liu2024apigen} introduced techniques for generating single-turn function calling data, while  \cite{su2025learn} explored methods for knowledge distillation in agent training. However, these approaches primarily focus on single-turn interactions, failing to capture the complexity of real-world agent usage, where multiple turns are often required. Other efforts like \cite{zeng2025boosting, yin2025magnetmultiturntoolusedata,chen2025facilitating}, while incorporating multi-turn aspects, lack human-agent interplay--crucial for realistic data generation. The verification and synthesis of high-quality multi-turn trajectories containing both linguistic diversity and grounded actions remains largely unsolved, creating a significant barrier to advancing agent capabilities.

To address these limitations, we introduce \textbf{\pipeline}, an agentic data synthesis pipeline for generating high-quality multi-turn agent data. It operates in two main steps: first, a data agent generates a detailed and verified task "blueprint", and second, this blueprint guides the generation of realistic multi-turn interactions through \textit{simulated agent-human interplay} (\autoref{sec:simulation}). The blueprint generation includes sampling relevant APIs, policies, domain data, and user personas to create grounded general tasks configurations, and using \textit{reverse task recombination} (\autoref{sec:recombination_taubench}) to enhance complexity. These blueprints are validated through format/execution checks and an LLM committee review using a reflection-based mechanism \cite{shinn2023reflexion}. Subsequently, the validated blueprint seeds a simulated interaction between a human LM and an agent (e.g., \texttt{gpt-4o}), producing a complete interaction trajectory with dialogue, actions, and environment feedback for training.

The main contributions of our work are summarized as follows:
\begin{itemize}[leftmargin=1.5em, itemsep=1pt, topsep=1pt, parsep=1pt]
\item We propose \textbf{\pipeline}, an agentic data synthesis pipeline that leverages environment execution feedback and a review committee to ensure the high-quality of generated multi-turn agent data.
\item We develop a two-phase framework that first creates detailed task blueprints with verifiable groundtruth actions, then transforms these blueprints into realistic multi-turn conversational agent trajectories with tool-usage through simulated human-agent interplay.
\item We train a series of models across multiple architectures and scales (Llama 3.1/3.2 and Qwen 2.5 at 1B to 70B parameters), demonstrating superior performance on two popular agentic benchmarks: $\tau$-bench and BFCL, surpassing many frontier models including \texttt{gpt-4o} (\autoref{fig:model_leaderboard}).
\item We open-source 5K high-quality synthetic data (\texttt{APIGen-MT-5k}) and trained models, i.e., the \modelseries{} series, to advance research in AI agent space.
\end{itemize}

\section{Related Work}
\label{sec:related}

\textbf{Tool-Use Agents.} Tool-use capabilities enhance LLMs by enabling interaction with external tools, extending their reasoning and functionality \cite{wölflein2025llm, qin2024tool,liu2024toolace}. Function-calling frameworks allow LLMs to parse queries, select tools, and interpret results, but often require predefined tools, limiting adaptability \cite{liu2024apigen,wang2024hammerbench}. Efforts were made to address this by creating reusable tools from scratch on the fly \cite{cai2023large}, built upon by
ToolMaker \cite{wölflein2025llm} which leverages tools from existing code repositories. Others compose workflows or learn from demonstrations \cite{qin2024tool, schick2023toolformer}. Recently, several works have adopted specialized approaches for agent training---critique-informed planning \cite{chen2025atlas}, fine-tuning on selective steps \cite{yang2025lighthouse}, teasing apart reasoning from format following (Agent-FLAN) \cite{chen2024agent}, and autonomously invoking tools without explicit post-training (ToRL) \cite{li2025torlscalingtoolintegratedrl}.

\textbf{Interactive Conversational Benchmarks.} Evaluating LLM agents in multi-turn settings requires specialized benchmarks. MultiChallenge \cite{sirdeshmukh2025multichallenge} and ToolDial \cite{shim2025tooldial} assess agents on context maintenance and tool-augmented dialogue. InterCode \cite{yang2024intercode} and CRMArena \cite{huang2025crmarenaunderstandingcapacityllm} evaluate iterative problem-solving and customer management. ToolSandbox \cite{lu2024toolsandbox} provides a stateful, interactive benchmark for tool use.
User simulations have become essential in these benchmarks, offering systematic, realistic interactions \cite{yao2024tau, lu2024toolsandbox, pan2025benchmarkstalkreevaluatingcode}. Our work complements these efforts by generating synthetic multi-turn conversations to train and evaluate agents in such realistic settings.

\textbf{Synthetic Data Generation.} The scarcity of high-quality training data drives synthetic data generation. Multi-agent frameworks like MAG-V \cite{sengupta2024mag}, AgentInstruct \cite{mitra2024agentinstruct}, MATRIX \cite{tang2024synthesizing}, and IntellAgent \cite{levi2025intellagent} create realistic datasets by simulating agent interactions. Other approaches utilize instruction composition \cite{hayati2024chain,chen2025facilitating}, intermediate graphs \cite{arcadinho2024automated} and multi-turn planning to produce complex dialogues \cite{zhang-etal-2024-probing}. Related to our effort in generating multi-turn training data, BUTTON \cite{chen2025facilitating} generates synthetic compositional instruction tuning data by combining 2-3 atomic tasks and conducting trajectory collection via a multi-agent setup. However, this involves construction of APIs based on the task generated and lacks systematic quality control and filtering during task composition limiting data verification. MAGNET \cite{yin2025magnetmultiturntoolusedata} proposed a graph-based method to generate function signature paths which are iteratively transformed to a sequence of queries and function calls.

While many of these prior approaches have been tested mainly on reasoning or single-turn interaction scenarios, our framework, \pipeline, advances this line of work, being applicable to any existing environment by generating high-quality multi-turn data for realistic agent-human interactions, focusing on reliable tool selection and parameter generation. By systematically preparing  the context, we first generate tasks adhering to any domain constraints and the corresponding executable groundtruth function calls in an agentic fashion with iterative refinement via feedback loops. Further, the simulated agent-human interplay mechanism allows us to generate verifiable long interaction trajectories.

\section{\pipeline ~Method for Synthesizing High-Quality Multi-Turn Data}
\label{sec:methods}

In this section, we present \pipeline, an agentic pipeline for generating multi-turn data through simulated agent-human interplay. We first formalize the multi-turn interaction problem and then describe our two-phase framework for generating high-quality, verifiable multi-turn data.

\subsection{Multi-Turn Interaction Problem Formulation}
\label{sec:problem_formulation}

Multi-turn interactions between an AI assistant and a human user present unique challenges that go beyond single-turn exchanges. We formalize this interaction as a Partially Observable Markov Decision Process (POMDP) defined by the tuple $(\mathcal{U}, \mathcal{S}, \mathcal{A}, \mathcal{O}, \mathcal{T}, \mathcal{R})$, where $\mathcal{U}$ represents the instruction space containing possible user intents; $\mathcal{S}$ denotes the state space of the environment and conversation history; $\mathcal{A} = \{tool\_call, response\}$ is the action space available to the assistant; $\mathcal{O} = \mathcal{O}_E \cup \mathcal{O}_H$ is the observation space comprising observations from the environment ($\mathcal{O}_E$) and response from the human ($\mathcal{O}_H$); $\mathcal{T}: \mathcal{S} \times \mathcal{A} \to \mathcal{S} \times \mathcal{O}$ is the transition function; and $\mathcal{R}$ is the reward function evaluating interaction success.
The AI assistant must engage in a multi-turn conversation with the human user to incrementally understand their intent $q \in \mathcal{U}$ and solve it through appropriate interactions with the environment while adhering to any domain rules. At turn $t$, the assistant predicts an action $a^t \in \mathcal{A}$ based on the interaction history and  understanding of $q$ thus far.
When $a^t$ is a $tool\_call$ compliant with the rules, it triggers a state transition $(s^t_E, tool\_call) \rightarrow (s^{t+1}_E, o_E)$, where $o_E \in \mathcal{O}_E$ is the tool output (typically in structured format like JSON). When $a^t$ is a $response$ to the human, it causes a state transition $(s^t_H, response) \rightarrow (s^{t+1}_H, o_H)$, where $o_H \in \mathcal{O}_H$ is the human's follow-up message. Importantly, the environment state $s^{t+1}_E$ remains latent to both the assistant and the human. The interaction completes when the human sends a terminating message or the maximum number of turns is reached. The reward $\mathcal{R}(\Delta\mathcal{S}_E, a)$ is calculated based on the cumulative state change in the environment $\Delta\mathcal{S}_E$ and the sequence of responses $a = \{a_i \mid a_i \in response \text{ to human}\}$ provided by the assistant throughout the episode. The assistant's objective is to maximize this reward.

\subsection{\pipeline~Framework Overview}
\label{sec:framework}

Generating high-quality multi-turn data that captures the complexities of agent-human interactions presents significant challenges. Directly synthesizing multi-turn conversations in one shot is difficult for two key reasons: (1) a single error or hallucination in any intermediate step can lead to complete failure, and (2) the content of each turn depends on previous function calls and their outputs, creating complex dependencies that are difficult to maintain consistently.

\begin{figure}[t]
\centering
\includegraphics[width=\linewidth]{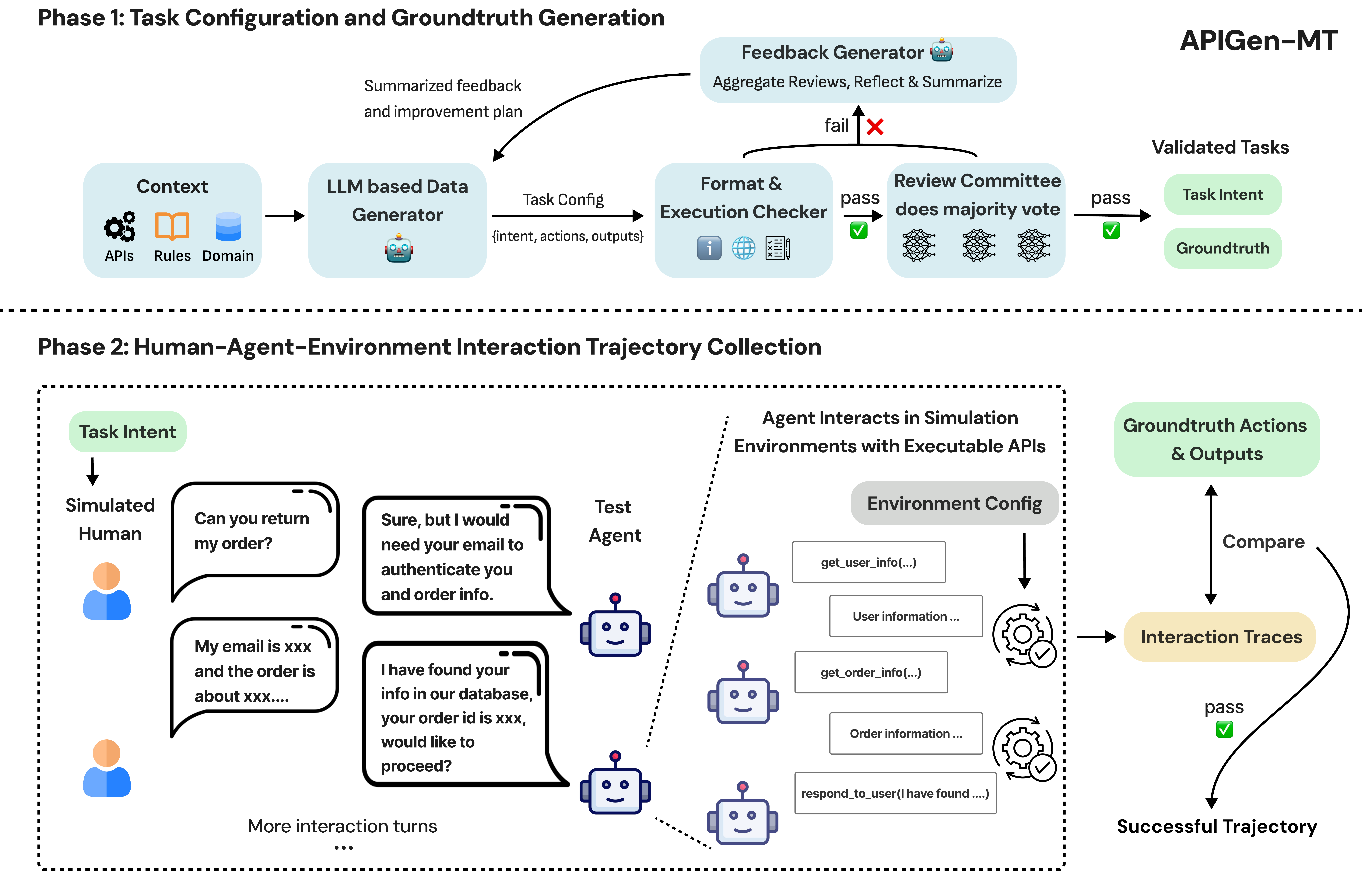}
\caption{\small{Overview of the \pipeline~framework. Phase 1 generates task configurations and groundtruth actions through an agentic process with feedback loops. Phase 2 collects human-agent-environment interaction trajectories by simulating realistic conversations between a human user and a test agent in an executable environment.}}
\label{fig:pipeline}
\end{figure}

To address these challenges, we introduce \pipeline, a two-phase framework for generating verifiable and diverse multi-turn data (\autoref{fig:pipeline}). Our approach extends the APIGen framework \cite{liu2024apigen} by adding an agentic feedback loop and simulated human-agent interplay to generate realistic multi-turn conversations.

The core insight of our approach is to separate the task generation process into two distinct phases: first creating a detailed "blueprint" of the task (Phase 1), and then using this blueprint to guide the generation of realistic multi-turn interactions that fill in the conversational details (Phase 2). This separation allows us to ensure both the correctness of the underlying task structure and the naturalness of the resulting conversations.

\subsubsection{Phase 1: Task Configuration and Groundtruth Generation}
\label{sec:phase1}
The initial phase of \pipeline~ focuses on systematically generating well-defined task configurations, each comprising a user instruction ($q$), a corresponding sequence of verifiable groundtruth actions ($a_{gt}$), and the expected final outputs ($o_{gt}$). This phase establishes a solid, verifiable foundation for each interaction scenario before the complexities of conversational dynamics are introduced. As depicted in \autoref{fig:pipeline}, this is achieved through an agentic workflow incorporating multi-stage validation and refinement loops. More specifically, it has the following steps:

\begin{enumerate}[itemsep=1pt, topsep=0pt, leftmargin=1.5em] %
    \item \textbf{Context Preparation}: Relevant information such as available APIs, domain-specific rules or policies, and reference data is assembled. This context grounds the subsequent generation step in the specific constraints and capabilities of the target environment.
    
    \item \textbf{LLM-based Data Generator}: An LLM utilizes the prepared context to propose initial task configurations. Each configuration consists of:
        \begin{itemize}[itemsep=0pt, topsep=0pt, leftmargin=1em]
            \item A detailed user instruction $q$ describing the high-level intent.
            \item A sequence of groundtruth actions $a_{gt}$ required to fulfill the intent.
            \item Expected final outputs $o_{gt}$ to be provided to the user.
        \end{itemize}

    \item \textbf{Format \& Execution Checker}: Proposed configurations undergo automated technical validation. This component performs multiple checks:
        \begin{itemize}[itemsep=0pt, topsep=0pt, leftmargin=1em]
            \item Verifies the structural correctness of generated actions (e.g., valid API call formats) and outputs.
            \item Confirms the executability of each action in $a_{gt}$ within a simulated target environment $E$ (checking API names, arguments, types).
        \end{itemize}
        
    \item \textbf{Review Committee}: Configurations passing rule-based checks proceed to semantic evaluation by a committee of multiple LLM reviewers. This committee assesses quality aspects like the coherence between $q$ and $a_{gt}$, completeness, and overall task sensibility. We use majority voting to achieve a more stable assessment.
        
    \item \textbf{Feedback Generation and Refinement}: If a task fails at either the validation (Step 3) or review (Step 4) stage, a Feedback Generator aggregates failure reasons and reviews, reflects upon them, and produces a summarized improvement plan. This plan guides the Data Generator (Step 2) in refining the task proposal in a subsequent iteration. Successfully validated tasks exit this loop.
\end{enumerate}

This agentic design with feedback loops is crucial for generating high-quality tasks efficiently. By incorporating reflection and improvement based on validation results, the system can learn from failures and progressively generate better tasks.

\subsubsection{Phase 2: Human-Agent-Environment Interaction Trajectory Collection}
\label{sec:phase2}

Building upon the validated task configurations ${q, a_{gt}, o_{gt}}$ from Phase 1, the second phase generates realistic multi-turn interaction data by simulating dynamic conversations between an LLM-based human user and a test agent  operating within an executable environment. Guided by the task instruction $q$ and often a specific persona, the simulated human naturally reveals information or sub-goals incrementally, while the agent interprets the evolving context, interacts with the environment via API calls when needed, and responds coherently. Importantly, the simulated user is unaware of the underlying environment and available APIs mimicking a real-world user. 

The simulation produces complete interaction trajectories that capture dialogue turns, agent actions, and environment responses. Each trajectory is validated by comparing its outcome against the groundtruth actions ($a_{gt}$) and expected outputs ($o_{gt}$) from Phase 1. Only those trajectories that verifiably achieve the task using both state-based and output-based checks are accepted into the dataset, ensuring that interactions are both dynamically plausible and grounded in a correct solution. 

This two-phase design offers several benefits. First, it provides verifiability by grounding interaction data in pre-validated task configurations. Second, it enhances realism by focusing the simulation on natural turn-by-turn dynamics without the simultaneous burden of task solution generation. Lastly, the modular approach isolates issues in task design from those in conversational modeling, facilitating debugging and scalability across diverse interaction patterns. 
In essence, by integrating agentic generation of verifiable task "blueprint" with realistic simulation of conversational dynamics, \pipeline~produces high-quality, multi-turn interaction data that balances structural correctness with the naturalness required for training agent models.

\section{A Case Study of \pipeline{} on $\tau$-bench}
\label{sec:implementation}

\begin{figure}[tbp]
\centering
\includegraphics[scale=2.5,width=1\linewidth]{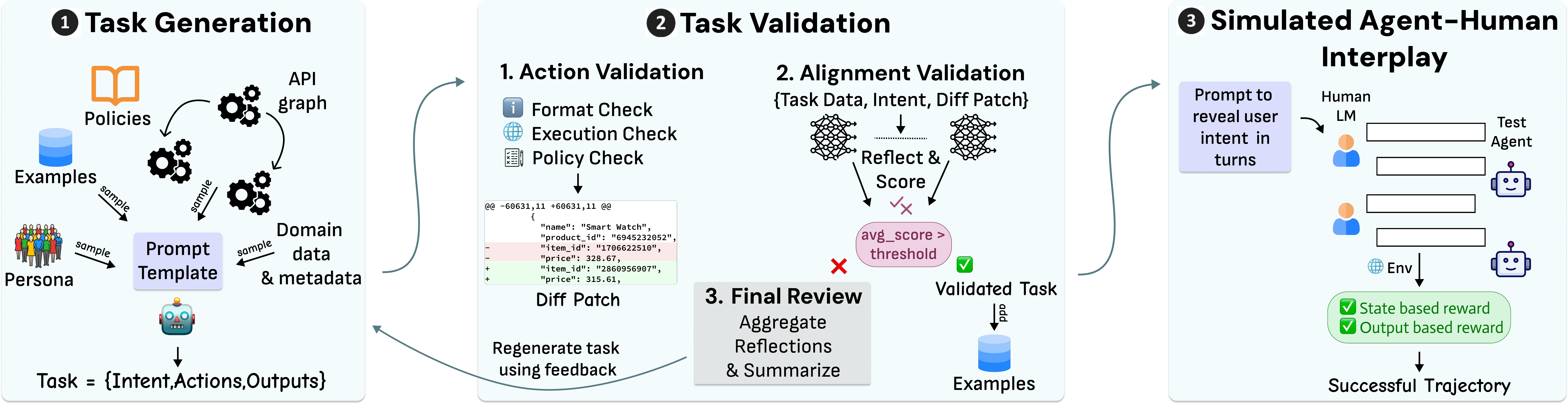}
\caption{\small{Realization of \pipeline{} framework for $\tau$-bench. We first generate realistic task instances by random walk down the API graph and sampling. Next the tasks are validated following a multi-stage pipeline. Instances which fail are sent back to the Generator to be refined based on the validation feedback. Finally, trajectories are generated by a simulated human user that interacts with a test agent by supplying the query details in a turn-wise manner. Trajectories which pass state- and output- based evaluations are collected.}}
\label{fig:preview}
\end{figure}

This section details the instantiation of the \pipeline~framework (\autoref{sec:framework}) with $\tau$-bench \cite{yao2024tau}. Generating high-quality, multi-turn interaction data with nuanced human-agent dynamics presents challenges, as direct conversation simulation often leads to inconsistencies or task deviations. Therefore, our two-phase approach addresses this by first synthesizing detail task configurations that define the user's high-level intent (\(q\)), groundtruth actions (\(a_{gt}\)), and the expected final outputs (\(o_{gt}\)). By establishing this verifiable "blueprint" first (Phase 1), we can then more reliably simulate the fine-grained, turn-by-turn interaction dynamics between a human and an agent within the executable environment (Phase 2), ensuring the collected trajectories are both realistic and grounded in a verifiable solution path. $\tau$-bench, with its realistic domains, executable APIs, and specific policies, provides an ideal testbed for this methodology. \autoref{fig:preview} illustrates this specific implementation.

\subsection{Phase 1 Implementation: Task Configuration Generation and Validation}
\label{sec:phase1_impl}

\subsubsection{API Dependency Graph and Context Samplers}

Generating realistic tasks for $\tau$-bench requires navigating its specific APIs, policies, and data structures. We implemented the following techniques for task generation and validation.

\paragraph{API Graph Modeling.} 
We model the available APIs in each $\tau$-bench domain as a directed graph, where nodes represent APIs and edges represent dependencies between them. An edge exists from API $A$ to API $B$ if $B$'s input arguments can depend on $A$'s output and the co-occurrence of this tool-call pair is permitted under domain policies. This graph-based approach enables us to generate realistic task sequences by performing random walks through the API dependency graph.

\paragraph{Specialized Context Samplers.}
To ensure task diversity, realism, and grounding, we utilize several domain-specific samplers that provide context to the LLM-based task generator.

\begin{itemize}[itemsep=1pt,topsep=0pt,leftmargin=1.5em]
    \item \textbf{API Sampler}: We distinguish between state-exploring (`read') APIs and state-changing (`write') APIs which can modify the environment states. The generator focuses on sampling the necessary 'write' APIs to form the core of $a_{gt}$, allowing flexibility in how `read' APIs might be used during the subsequent interaction phase.  This approach encourages exploration while ensuring that specific state-changing actions are included in the groundtruth.
    
    \item \textbf{Policy Sampler}: For each $\tau$-bench domain, we extract and sample from the domain-specific policies and rules. These policies are incorporated into the task generation process to ensure compliance of real-world use cases. Task complexity is influenced by the number of 'write' calls and the associated policy constraints.

    \item \textbf{Domain Data Sampler}: To ground tasks in realistic domain data without exceeding context limits, we sample domain-specific data with additional metadata (e.g., cost, time, attributes). This metadata enhances coverage and enables more creative and diverse task scenarios.
    
    \item \textbf{Persona Sampler}: We incorporate user persona descriptions from PersonaHub \cite{ge2024scalingsyntheticdatacreation} to inform the user intent $q$ and inject realistic human qualities and situational context, enhancing diversity for subsequent Phase 2 human-agent interaction simulation.
    \item \textbf{Example Sampler}: We provide few-shot examples of well-formed tasks relevant to the sampled APIs, guiding the generator on structure and format.
\end{itemize}

For each task generation iteration, we randomly vary the sampling frequency for each sampler to enhance diversity and prevent repetitive scenarios. The sampled information is compiled into a prompt instructing the LLM generator to produce a \texttt{<thought>} (its reasoning process), the user \texttt{<instruction>} ($q$), the corresponding groundtruth \texttt{<actions>} ($a_{gt}$), and the expected final \texttt{<outputs>} ($o_{gt}$).

\subsubsection{Multi-Stage Validation for $\tau$-bench}
\label{sec:validation_taubench}

We implement a rigorous three-stage validation process for the $\tau$-bench environment:

\paragraph{Stage 1: Action Validation.}
\begin{itemize}[itemsep=1pt,topsep=0pt,leftmargin=1.5em]
    \item \textbf{Format Check}: Verifies the presence and basic structure of required task components (\texttt{<thought>}, \texttt{<instruction>}, \texttt{<actions>}, \texttt{<outputs>}) and ensures all tool calls in \texttt{<actions>} are valid JSON and outputs in \texttt{<outputs>} are strings.
    \item \textbf{Execution Check}: Simulates each action in $a_{gt}$ within the  $\tau$-bench~environment, validating API names, argument names, and data types. The cumulative effect on the environment state ($\Delta\mathcal{S}_E$) is captured as a \texttt{diff\_patch}, similar to \texttt{git diff}.
    \item \textbf{Policy Compliance Check}: Leverages the executable nature of  $\tau$-bench~by translating domain policies into Python unit tests. These tests run against the simulated execution trace of $a_{gt}$ to detect violations, especially those arising from interactions between multiple actions (e.g., action B is invalid given the state change caused by prior action A). Failures yield detailed feedback on the specific policy violation.
\end{itemize}

\paragraph{Stage 2: Alignment Validation.}
Tasks successfully passing Stage 1's action validation are then assessed for semantic alignment. Specifically, we evaluate whether the groundtruth actions ($a_{gt}$), as reflected by their environmental effects summarized in the \texttt{diff\_patch}, accurately and comprehensively fulfill the user's intent expressed in the instruction ($q$). To mitigate the potential biases and inconsistencies of a single evaluator, we employ a committee of diverse LLM judges \cite{zhang2024diversity,bi2024forest}. These judges review each task based on a systematic rubric with metrics such as Correctness, Completeness, Satisfaction, and Creativity (refer \autoref{fig:validation_prompt} in \autoref{appx:prompts} for details).

Each judge provides scores and qualitative feedback. We utilize a majority voting strategy across the committee's judgments to determine the final assessment for each metric and the overall task quality. This approach yields more stable and reliable evaluation results compared to single-judge assessments.

\paragraph{Stage 3: Final Semantic Review \& Refinement.}
Based on the aggregated scores from the committee (determined via majority voting), tasks achieving an average score above a predefined threshold are accepted and added to the pool of validated task configurations. Tasks that fail this review trigger the feedback loop mechanism. Consolidated feedback, summarizing the points raised by the committee majority, is sent back to the LLM task generator. This initiates a reflection process \cite{shinn2023reflexion}, guiding the generator to revise the task in the subsequent iteration to address the identified shortcomings.

\subsubsection{Reverse Task Recombination for Complex Task Construction}
\label{sec:recombination_taubench}
While the iterative refinement process improves task quality and efficiency, directly generating complex, long-horizon tasks involving multiple steps remains challenging. Validation failures can occur due to subtle policy conflicts or difficulties in ensuring perfect alignment across many steps. To overcome this and systematically construct more complicated scenarios, we implement \textit{Reverse Task Recombination}, a technique that leverages the principle of compositionality \cite{chen2025facilitating,hayati2024chain}, similar to modular design in software engineering. The core idea is to build complex tasks from simpler, independently validated "building blocks":
\begin{enumerate}[itemsep=1pt,topsep=0pt,leftmargin=1.5em]
    \item \textbf{Select Validated Tasks}: Identify multiple simpler tasks ($T_1, T_2, ...$) that have successfully passed all validation stages (Stages 1-3) and are associated with the same user persona.
    \item \textbf{Concatenate Components}: Combine their respective groundtruth actions ($a_{combined} = a_{gt,1} \circ a_{gt,2} \circ ...$) and expected outputs ($o_{combined} = o_{gt,1} \oplus o_{gt,2} \oplus ...$, where $\circ$ denotes action sequence concatenation and $\oplus$ denotes output aggregation).
    \item \textbf{Re-Check Policy Compliance}: Rerun the Policy Check on $a_{combined}$ to ensure that the cumulative action sequence remains logically sound and adheres to the domain rules as combinations could cause conflicting actions to appear together, for e.g., returning and canceling the same order. 
    \item \textbf{Synthesize Combined Instruction}: Instruct the LLM generator to create a new, coherent, overarching user instruction ($q_{combined}$) that logically integrates the goals and steps represented by $a_{combined}$ and $o_{combined}$. This new instruction should frame the combined actions as a single, more complex user request.
    \item \textbf{Re-Validate Semantics}: Submit the newly formed complex task $T_{combined} = \{q_{combined}, a_{combined}, o_{combined}\}$ for validation starting from Stage 2 (Alignment Validation). Stage 1 (Action Validation) can be safely skipped for $a_{combined}$ because each constituent action sequence ($a_{gt,1}, a_{gt,2}, ...$) has already been individually checked for format and execution within its original context, and policy compliance in the current context. Stage 3 (Final Semantic Review) proceeds based on the outcome of Stage 2 for the combined task.
\end{enumerate}
This method allows for the scalable generation of complex, multi-step tasks with greater reliability, as it builds upon verified components while focusing the validation effort on the semantic coherence and alignment of the combined whole.

\subsection{Phase 2: Simulated Human-Agent Interplay and Trajectory Collection}
\label{sec:simulation} %

Building on the verified tasks from Phase 1—which include a detailed user intent $q$, groundtruth actions $a_{gt}$, and expected outputs $o_{gt}$—we simulate multi-turn interaction trajectories between an agent ($A$) and a human user ($H$) modeled by an LLM. Guided by the instruction $q$ and an associated persona, the simulated human incrementally reveals task details to mimic realistic interactions. The agent, instantiated as \texttt{gpt-4o} with its function-calling mode, interprets the evolving intent and executes the necessary actions to complete the task.

\textbf{Trajectory Collection.  } We employ rejection sampling to ensure that only trajectories achieving the task goal ($r=1$) are retained. Success is determined by comparing the final environment state to $a_{gt}$ and the agent’s final responses to $o_{gt}$. For enhanced data coverage, each task is attempted up to three times, and the union of all unique successful trajectories is compiled into an offline dataset suitable for downstream applications such as behavioral cloning.

\textbf{Stabilizing Simulated Human.  } A critical challenge in this phase is maintaining the stability and fidelity of the simulated human. Over multiple conversational turns, the human LLM may drift from the original instruction or be unduly influenced by the agent's responses \cite{park2023generative}, introducing variability that hinders reliable evaluation \cite{yao2024tau}. To address this, we adopt a Best-of-N (N=4) sampling strategy in combination with a self-critique mechanism for the human LLM's responses (see \autoref{fig:bon_prompt} in \autoref{appx:prompts} for details), allowing it to adhere to the task instruction more accurately and not be mislead by the test agent responses. Its effectiveness was validated on the $\tau$-bench test set, where improved consistency in agent performance evaluation across multiple trials was observed (\autoref{tab:stability_results}).

\subsection{Data Collection \& Statistics}
\label{sec:synthetic_stats}

\textbf{Data Collection Procedure. } 
We source APIs implemented as Python functions from $\tau$-bench. Among these, we have 15 `read' and  13 `write' APIs across both domains. $\tau$-bench is accompanied with detailed policies and domain rules in two settings - Retail and Airline which we use as guideline policies. We utilize \texttt{gpt-4o} and DeepSeek V3 models in the task generation, validation and agent-human interplay stages to collect training data. The prompts used in every stage are provided in  \autoref{appx:prompts}. We set the maximum number of reflection-based feedback  turns to 3 for retail and 5 for airline respectively. 

\begin{figure}[h] %
    \centering %
    \begin{minipage}{0.48\textwidth} %
        \centering
        \small
        \begin{tabular}{lr} %
        \toprule 
        \textbf{Metric} & \textbf{Value} \\ %
        \midrule
        Task Config. S.R. (Phase 1) & 70\% \\
        Task Config. S.R. w/o Agentic Feedback & 28\% \\
        Trajectory Sim. S.R. (Phase 2) & 67\% \\
        \midrule %
        Min. Turns per Trajectory & 1 \\
        Max. Turns per Trajectory & 29 \\
        Avg. Tool Calls per Trajectory & 7 \\
        Avg. User Turns per Trajectory & 6 \\
        \bottomrule
    \end{tabular}
    \caption{\small Statistics for the dataset generated using \textbf{\pipeline}. Success rates (S.R.) are reported for the task configuration (w. and w/o agentic feedback in Phase 1) and trajectory simulation (Phase 2) stages.} 
    \label{table:data_stats}
    \end{minipage}\hfill %
    \begin{minipage}{0.42\textwidth} %
        \centering
        \includegraphics[width=\linewidth]{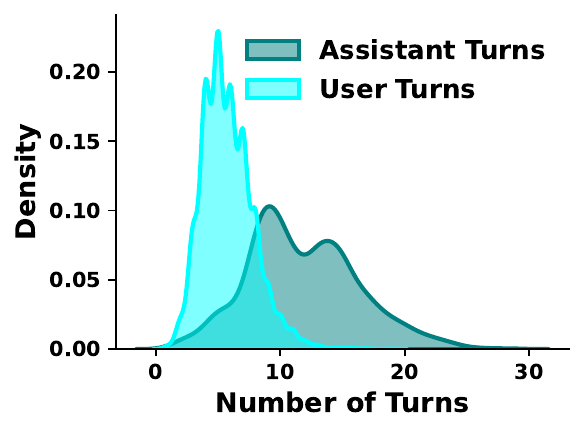} %
        \caption{\small{Density distribution of assistant and user turns in collected trajectories.}}
        \label{fig:data_stats}
    \end{minipage}

\end{figure}

\textbf{Statistics. } 
A summary of the data collection is shown in \autoref{table:data_stats}. \autoref{fig:data_stats} shows that we can efficiently collect long trajectories requiring a strong model like \texttt{gpt-4o} to take an average 12 turns to complete the task using \pipeline. Our agentic pipeline involving review committee and iterative refinement via reflection provides a \textbf{2.5x} boost to the task collection success rate to attain 70\%.

Our implementation demonstrates that the \pipeline~framework can successfully generate high-quality multi-turn data for complex domains with strict policy constraints. The two-phase approach with agentic feedback loops and simulated human-agent interplay proves effective in creating diverse, realistic, and verifiable datasets for training and evaluating conversational agents.

\section{Experiments}
\label{sec:experiment}

\subsection{Experimental Setup}
\textbf{Training Details.  } We perform filtered Behavioral Cloning (BC) using the collected trajectories with Llama 3.1/3.2 Instruct models \cite{dubey2024llama} and Qwen 2.5 Instruct models \cite{qwen2025qwen25technicalreport}. The collected trajectories are split at every assistant response and we train to predict only the assistant response tokens by masking the prompt and other messages.
To enhance the dataset diversity, we also jointly train our \modelseries{} models with function-calling data from \cite{liu2024apigen} and other domains of agentic data from \cite{zhang2024xlam,zhang2025actionstudio}.
We utilize the LLama-Factory library \cite{zheng2024llamafactory} and perform full-finetuning using DeepSpeed ZeRO \cite{10.1145/3394486.3406703} stage 3, Flash Attention 2 \cite{dao2023flashattention2fasterattentionbetter} in bfloat16 precision with AdamW optimizer \cite{loshchilov2017decoupled} and train for at most 3 epochs on a NVIDIA H200 node.

\textbf{Benchmarks.  } We evaluate on two challenging benchmarks designed specifically for assessing agent capabilities -- \textbf{(1) BFCL v3} \cite{berkeley-function-calling-leaderboard}, a leading benchmark for tool-use evaluation, specifically designed to assess LLMs' function calling capabilities and \textbf{(2) $\boldsymbol{\tau}$-bench} \cite{yao2024tau}, a comprehensive benchmark for evaluating AI agents in realistic scenarios. More details are in \autoref{appx:benchmarks}. Both are particularly well-suited for evaluating the effectiveness of our \pipeline{} approach, as they focus on multi-turn interactions and tool use capabilities, which are central to our data generation methodology.

\subsection{Experiment Results}
We compare the performance of our trained models (\modelseries{}) against state-of-the-art proprietary models such as \texttt{gpt} models (\texttt{o1}, \texttt{gpt-4o}); \texttt{claude} models (\texttt{claude-3.5-haiku}, \texttt{claude-3.5-sonnet}, \texttt{claude-3.5-sonnet (new)}, and \texttt{claude-3.7-sonnet}), and open-source LLMs including DeepSeek v3, and the baselines Llama 70B and Qwen 32B.

\paragraph{BFCL v3 Results.} On the BFCL v3 benchmark, our models demonstrate exceptional performance. As shown in \autoref{table:bfcl-comparison}, \modelname{} and \texttt{xLAM-2-32b-fc-r} achieve the top 2 positions on the leaderboard with overall accuracies of 78.19\% and 75.83\% respectively, surpassing all proprietary and open-source models. The most striking advantage appears in multi-turn scenarios, where our models excel across all parameter scales. \modelname{} achieves 75.12\% multi-turn accuracy, while our smaller models show remarkable capabilities with \texttt{xLAM-2-8b-fc-r} at 69.25\%, \texttt{xLAM-2-3b-fc-r} at 56.00\%, and even \texttt{xLAM-2-1b-fc-r} at 43.12\% - all substantially outperforming o1 (36\%) and \texttt{gpt-4o} in function-calling mode (41\%). Additionally, our models demonstrate strong hallucination detection, with \texttt{xLAM-2-3b-fc-r} achieving 94.44\% on relevance detection, matching the best score in this category.\\

\begin{table}[t]
\centering
\caption{\small Performance of different models on BFCL leaderboard (as of date 04/03/2025). The rank is based on the overall accuracy, which is a weighted average of different evaluation categories. ``FC" stands for function-calling mode in contrast to using a customized ``prompt" to extract the function calls. See the benchmark \cite{berkeley-function-calling-leaderboard} for details.}
\label{table:bfcl-comparison}
\resizebox{1.\linewidth}{!}{
\begin{tabular}{|c|c|c|cccccc|}
\hline
{\color[HTML]{000000} }                       & {\color[HTML]{000000} }                              & {\color[HTML]{000000} }                                    & \multicolumn{3}{c|}{}                                                                                                                                              & \multicolumn{1}{c|}{{\color[HTML]{000000} }}                                       & \multicolumn{2}{c|}{}                                                                                     \\
{\color[HTML]{000000} }                       & {\color[HTML]{000000} }                              & {\color[HTML]{000000} }                                    & \multicolumn{3}{c|}{\multirow{-2}{*}{Single-Turn}}                                                                                                                 & \multicolumn{1}{c|}{\multirow{-2}{*}{{\color[HTML]{000000} Multi-Turn}}}           & \multicolumn{2}{c|}{\multirow{-2}{*}{Hallucination}}                                                      \\ \cline{4-9} 
{\color[HTML]{000000} }                       & {\color[HTML]{000000} }                              & {\color[HTML]{000000} }                                    &                                       &                                       & \multicolumn{1}{c|}{}                                                              & \multicolumn{1}{c|}{{\color[HTML]{000000} }}                                       & {\color[HTML]{000000} }                            & {\color[HTML]{000000} }                              \\
\multirow{-4}{*}{{\color[HTML]{000000} Rank}} & \multirow{-4}{*}{{\color[HTML]{000000} Overall Acc}} & \multirow{-4}{*}{{\color[HTML]{000000} Model}}             & \multirow{-2}{*}{Non-live (AST)}      & \multirow{-2}{*}{Non-live (Exec)}     & \multicolumn{1}{c|}{\multirow{-2}{*}{Live (AST)}}                                  & \multicolumn{1}{c|}{\multirow{-2}{*}{{\color[HTML]{000000} Overall Acc}}}          & \multirow{-2}{*}{{\color[HTML]{000000} Relevance}} & \multirow{-2}{*}{{\color[HTML]{000000} Irrelevance}} \\ \hline 
\rowcolor[HTML]{C6DDFE} 
{\color[HTML]{000000} 1}                      & {\color[HTML]{000000} \textbf{78.19}}                & {\color[HTML]{000000} { \textbf{xLAM-2-70b-fc-r (FC)}}} & {\color[HTML]{000000} \textbf{88.48}} & {\color[HTML]{000000} \textbf{85.98}} & \multicolumn{1}{c|}{\cellcolor[HTML]{C6DDFE}{\color[HTML]{000000} \textbf{72.63}}} & \multicolumn{1}{c|}{\cellcolor[HTML]{C6DDFE}{\color[HTML]{000000} \textbf{75.12}}} & {\color[HTML]{000000} \textbf{66.67}}              & {\color[HTML]{000000} \textbf{78.74}}                \\
\rowcolor[HTML]{DAE8FC} 
{\color[HTML]{000000} \textbf{2}}             & {\color[HTML]{000000} \textbf{75.83}}                & {\color[HTML]{000000} { \textbf{xLAM-2-32b-fc-r (FC)}}} & {\color[HTML]{000000} \textbf{89.50}} & {\color[HTML]{000000} \textbf{86.48}} & \multicolumn{1}{c|}{\cellcolor[HTML]{DAE8FC}{\color[HTML]{000000} \textbf{73.79}}} & \multicolumn{1}{c|}{\cellcolor[HTML]{DAE8FC}{\color[HTML]{000000} \textbf{66.38}}} & {\color[HTML]{000000} \textbf{83.33}}              & {\color[HTML]{000000} \textbf{76.25}}                \\
{\color[HTML]{000000} 3}                      & {\color[HTML]{000000} 74.31}                         & {\color[HTML]{000000} watt-tool-70b (FC)}                  & {\color[HTML]{000000} 84.06}          & {\color[HTML]{000000} 89.39}          & \multicolumn{1}{c|}{{\color[HTML]{000000} 77.74}}                                  & \multicolumn{1}{c|}{{\color[HTML]{000000} 58.75}}                                  & {\color[HTML]{000000} 94.44}                       & {\color[HTML]{000000} 76.32}                         \\
\rowcolor[HTML]{E8F2FF} 
{\color[HTML]{000000} \textbf{4}}             & {\color[HTML]{000000} \textbf{72.83}}                & {\color[HTML]{000000} { \textbf{xLAM-2-8b-fc-r (FC)}}}  & {\color[HTML]{000000} \textbf{84.35}} & {\color[HTML]{000000} \textbf{85.59}} & \multicolumn{1}{c|}{\cellcolor[HTML]{E8F2FF}{\color[HTML]{000000} \textbf{66.73}}} & \multicolumn{1}{c|}{\cellcolor[HTML]{E8F2FF}{\color[HTML]{000000} \textbf{69.25}}} & {\color[HTML]{000000} \textbf{83.33}}              & {\color[HTML]{000000} \textbf{64.11}}                \\
{\color[HTML]{000000} 5}                      & {\color[HTML]{000000} 72.08}                         & {\color[HTML]{000000} GPT-4o-2024-11-20 (Prompt)}          & {\color[HTML]{000000} 88.1}           & {\color[HTML]{000000} 89.38}          & \multicolumn{1}{c|}{{\color[HTML]{000000} 79.83}}                                  & \multicolumn{1}{c|}{{\color[HTML]{000000} 47.62}}                                  & {\color[HTML]{000000} 83.33}                       & {\color[HTML]{000000} 83.76}                         \\
{\color[HTML]{000000} 6}                      & {\color[HTML]{000000} 69.94}                         & {\color[HTML]{000000} GPT-4.5-Preview-02-27 (FC)}          & {\color[HTML]{000000} 86.12}          & {\color[HTML]{000000} 83.98}          & \multicolumn{1}{c|}{{\color[HTML]{000000} 79.34}}                                  & \multicolumn{1}{c|}{{\color[HTML]{000000} 45.25}}                                  & {\color[HTML]{000000} 66.67}                       & {\color[HTML]{000000} 83.64}                         \\
{\color[HTML]{000000} 7}                      & {\color[HTML]{000000} 69.58}                         & {\color[HTML]{000000} GPT-4o-2024-11-20 (FC)}              & {\color[HTML]{000000} 87.42}          & {\color[HTML]{000000} 89.2}           & \multicolumn{1}{c|}{{\color[HTML]{000000} 79.65}}                                  & \multicolumn{1}{c|}{{\color[HTML]{000000} 41}}                                     & {\color[HTML]{000000} 83.33}                       & {\color[HTML]{000000} 83.15}                         \\
{\color[HTML]{000000} 8}                      & {\color[HTML]{000000} 68.39}                         & {\color[HTML]{000000} ToolACE-2-8B (FC)}                   & {\color[HTML]{000000} 87.58}          & {\color[HTML]{000000} 87.11}          & \multicolumn{1}{c|}{{\color[HTML]{000000} 80.05}}                                  & \multicolumn{1}{c|}{{\color[HTML]{000000} 36.88}}                                  & {\color[HTML]{000000} 72.22}                       & {\color[HTML]{000000} 90.11}                         \\
{\color[HTML]{000000} 9}                      & {\color[HTML]{000000} 67.98}                         & {\color[HTML]{000000} watt-tool-8B (FC)}                   & {\color[HTML]{000000} 86.56}          & {\color[HTML]{000000} 89.34}          & \multicolumn{1}{c|}{{\color[HTML]{000000} 76.5}}                                   & \multicolumn{1}{c|}{{\color[HTML]{000000} 39.12}}                                  & {\color[HTML]{000000} 83.33}                       & {\color[HTML]{000000} 83.15}                         \\
{\color[HTML]{000000} 10}                     & {\color[HTML]{000000} 67.88}                         & {\color[HTML]{000000} GPT-4-2024-04-09 (FC)}               & {\color[HTML]{000000} 84.73}          & {\color[HTML]{000000} 85.21}          & \multicolumn{1}{c|}{{\color[HTML]{000000} 80.5}}                                   & \multicolumn{1}{c|}{{\color[HTML]{000000} 38.12}}                                  & {\color[HTML]{000000} 72.22}                       & {\color[HTML]{000000} 83.81}                         \\
{\color[HTML]{000000} 11}                     & {\color[HTML]{000000} 67.87}                         & {\color[HTML]{000000} o1-2024-12-17 (Prompt)}              & {\color[HTML]{000000} 85.67}          & {\color[HTML]{000000} 87.45}          & \multicolumn{1}{c|}{{\color[HTML]{000000} 80.63}}                                  & \multicolumn{1}{c|}{{\color[HTML]{000000} 36}}                                     & {\color[HTML]{000000} 72.22}                       & {\color[HTML]{000000} 87.78}                         \\
{\color[HTML]{000000} 12}                     & {\color[HTML]{000000} 67.72}                         & {\color[HTML]{000000} BitAgent-8B}                         & {\color[HTML]{000000} 86.92}          & {\color[HTML]{000000} 89.52}          & \multicolumn{1}{c|}{{\color[HTML]{000000} 76.14}}                                  & \multicolumn{1}{c|}{{\color[HTML]{000000} 38.5}}                                   & {\color[HTML]{000000} 83.33}                       & {\color[HTML]{000000} 82.38}                         \\
{\color[HTML]{000000} 13}                     & {\color[HTML]{000000} 65.12}                         & {\color[HTML]{000000} o3-mini-25-01-31 (Prompt)}           & {\color[HTML]{000000} 86.15}          & {\color[HTML]{000000} 89.46}          & \multicolumn{1}{c|}{{\color[HTML]{000000} 79.08}}                                  & \multicolumn{1}{c|}{{\color[HTML]{000000} 28.75}}                                  & {\color[HTML]{000000} 72.22}                       & {\color[HTML]{000000} 82.96}                         \\
\rowcolor[HTML]{ECF4FF} 
{\color[HTML]{000000} \textbf{14}}            & {\color[HTML]{000000} \textbf{65.11}}                & {\color[HTML]{000000} { \textbf{xLAM-2-3b-fc-r (FC)}}}  & {\color[HTML]{000000} \textbf{82.94}} & {\color[HTML]{000000} \textbf{81.88}} & \multicolumn{1}{c|}{\cellcolor[HTML]{ECF4FF}{\color[HTML]{000000} \textbf{58.69}}} & \multicolumn{1}{c|}{\cellcolor[HTML]{ECF4FF}{\color[HTML]{000000} \textbf{56.00}}} & {\color[HTML]{000000} \textbf{94.44}}              & {\color[HTML]{000000} \textbf{57.94}}                \\
{\color[HTML]{000000} 15}                     & {\color[HTML]{000000} 64.1}                          & {\color[HTML]{000000} CoALM-405B}                          & {\color[HTML]{000000} 90.58}          & {\color[HTML]{000000} 89.07}          & \multicolumn{1}{c|}{{\color[HTML]{000000} 74.5}}                                   & \multicolumn{1}{c|}{{\color[HTML]{000000} 28.75}}                                  & {\color[HTML]{000000} 100}                         & {\color[HTML]{000000} 71.79}                         \\
{\color[HTML]{000000} 16}                     & {\color[HTML]{000000} 64.1}                          & {\color[HTML]{000000} GPT-4o-mini-24-07-18 (FC)}           & {\color[HTML]{000000} 85.21}          & {\color[HTML]{000000} 83.57}          & \multicolumn{1}{c|}{{\color[HTML]{000000} 74.41}}                                  & \multicolumn{1}{c|}{{\color[HTML]{000000} 34.12}}                                  & {\color[HTML]{000000} 83.33}                       & {\color[HTML]{000000} 74.75}                         \\ \hline
{\color[HTML]{000000} }                       & {\color[HTML]{000000} }                              & {\color[HTML]{000000} }                                    & \multicolumn{6}{c|}{{\color[HTML]{000000} }}                                                                                                                                                                                                                                                                                                                        \\
\multirow{-2}{*}{{\color[HTML]{000000} …}}    & \multirow{-2}{*}{{\color[HTML]{000000} …}}           & \multirow{-2}{*}{{\color[HTML]{000000} ...}}               & \multicolumn{6}{c|}{\multirow{-2}{*}{{\color[HTML]{000000} …}}}                                                                                                                                                                                                                                                                                                     \\ \hline
{\color[HTML]{000000} 34}                     & {\color[HTML]{000000} 58.93}                         & {\color[HTML]{000000} Gemini-2-Flash-Thinking}             & {\color[HTML]{000000} 87.4}           & {\color[HTML]{000000} 87.07}          & \multicolumn{1}{c|}{{\color[HTML]{000000} 75.97}}                                  & \multicolumn{1}{c|}{{\color[HTML]{000000} 14.5}}                                   & {\color[HTML]{000000} 77.78}                       & {\color[HTML]{000000} 72.75}                         \\
{\color[HTML]{000000} 35}                     & {\color[HTML]{000000} 58.9}                          & {\color[HTML]{000000} Qwen2.5-14B-Instruct (FC)}           & {\color[HTML]{000000} 85.42}          & {\color[HTML]{000000} 84.86}          & \multicolumn{1}{c|}{{\color[HTML]{000000} 76.68}}                                  & \multicolumn{1}{c|}{{\color[HTML]{000000} 15.88}}                                  & {\color[HTML]{000000} 55.56}                       & {\color[HTML]{000000} 77.69}                         \\
\rowcolor[HTML]{F0F6FF} 
{\color[HTML]{000000} \textbf{36}}            & {\color[HTML]{000000} \textbf{58.90}}                & {\color[HTML]{000000} { \textbf{xLAM-2-1b-fc-r (FC)}}}  & {\color[HTML]{000000} \textbf{76.23}} & {\color[HTML]{000000} \textbf{74.86}} & \multicolumn{1}{c|}{\cellcolor[HTML]{F0F6FF}{\color[HTML]{000000} \textbf{59.88}}} & \multicolumn{1}{c|}{\cellcolor[HTML]{F0F6FF}{\color[HTML]{000000} \textbf{43.12}}} & {\color[HTML]{000000} \textbf{88.89}}              & {\color[HTML]{000000} \textbf{56.87}}                \\
{\color[HTML]{000000} 37}                     & {\color[HTML]{000000} 58.55}                         & {\color[HTML]{000000} DeepSeek-V3 (FC)}                    & {\color[HTML]{000000} 89.17}          & {\color[HTML]{000000} 92.32}          & \multicolumn{1}{c|}{{\color[HTML]{000000} 68.41}}                                  & \multicolumn{1}{c|}{{\color[HTML]{000000} 18.62}}                                  & {\color[HTML]{000000} 88.89}                       & {\color[HTML]{000000} 59.36}                         \\
{\color[HTML]{000000} 38}                     & {\color[HTML]{000000} 58.45}                         & {\color[HTML]{000000} mistral-large-2407 (FC)}             & {\color[HTML]{000000} 86.81}          & {\color[HTML]{000000} 84.38}          & \multicolumn{1}{c|}{{\color[HTML]{000000} 69.88}}                                  & \multicolumn{1}{c|}{{\color[HTML]{000000} 23.75}}                                  & {\color[HTML]{000000} 72.22}                       & {\color[HTML]{000000} 52.85}                         \\
{\color[HTML]{000000} 39}                     & {\color[HTML]{000000} 58.42}                         & {\color[HTML]{000000} ToolACE-8B (FC)}                     & {\color[HTML]{000000} 87.54}          & {\color[HTML]{000000} 89.21}          & \multicolumn{1}{c|}{{\color[HTML]{000000} 78.59}}                                  & \multicolumn{1}{c|}{{\color[HTML]{000000} 7.75}}                                   & {\color[HTML]{000000} 83.33}                       & {\color[HTML]{000000} 87.88}                         \\ \hline
\end{tabular}
}
\end{table}

\paragraph{$\boldsymbol{\tau}$-bench Results.} \autoref{tab:main_results} presents results under the default \textit{naive} user setting on $\tau$-bench. Our \modelname{} model achieves a 56.2\% success rate, outperforming Llama 3.1 70B Instruct (38.2\%), DeepSeek v3 (40.6\%), and even proprietary models like GPT-4o (52.9\%), while approaching more recent models like Claude 3.5 Sonnet (60.1\%). Notably, our smaller variants like \texttt{xLAM-2-32b-fc-r} (54.6\%) and \texttt{xLAM-2-8b-fc-r} (46.7\%) surpass larger baselines, demonstrating that our synthetic data approach enables efficient knowledge transfer and strong performance with fewer parameters.

\begin{table}[h!]
\caption{\small Success Rate ($pass@1$) of various open-source and proprietary models on the Retail and Airline settings of $\tau$-bench (averaged across at least 5 trials). The \modelseries~ models are trained on the data generated using \pipeline . Overall indicates the average score across both domains. $^1$ indicates results from \cite{apt}; $^2$ indicates results from \cite{claude3.5}; $^3$ indicate results from \cite{claude3.7}; $^4$ indicates from \cite{claude_think}. \textbf{Note.} We evaluate only with the benchmark's \textit{think} tool and no prompt optimizations.}
\label{tab:main_results}
\centering
\small
\renewcommand\arraystretch{1.1}  %
\begin{tabular}{l|c|c|c}
\toprule
\textbf{Model} & \textbf{$\boldsymbol{\tau}$-Retail}
 & \textbf{$\boldsymbol{\tau}$-Airline} & \textbf{Overall}
 \\
\midrule
\multicolumn{4}{c}{\textit{Open-Source Models}} \\ \midrule
Qwen 2.5 32B Instruct & 24.4 & 25.0 & \gca{24.7} \\
Llama 3.1 70B Instruct & 50.4 & 26.0 & \gca{38.2} \\
DeepSeek v3$^1$	 & 58.3	& 22.8 & \gca{40.6} \\
\textbf{xLAM-2-70b-fc-r} & 67.1 & 45.2 & \gca{56.2} \\
\textbf{xLAM-2-32b-fc-r} & 64.3 & 45.0 & \gca{54.6} \\
\textbf{xLAM-2-8b-fc-r} & 58.2	& 35.2 & \gca{46.7} \\
\textbf{xLAM-2-3b-fc-r} & 44.4 & 32.0 & \gca{38.2} \\
\textbf{xLAM-2-1b-fc-r} & 22.5 & 21.0 & \gca{21.8} \\
\midrule
\multicolumn{4}{c}{\textit{Proprietary Models}} \\ \midrule
Gemini 1.5 pro$^1$ &	54.9 &	25.2 & \gca{40.1} \\
\texttt{gpt-4o-2024-11-20} & 62.8 & 43.0 & \gca{52.9} \\
\texttt{o1}$^3$	& 73.5	& 54.2 & \gca{63.9} \\
Claude 3.5 Haiku$^2$&	51.0	& 22.8 & \gca{36.9} \\
Calude 3.5 Sonnet$^2$ &	62.6 & 36.0 & \gca{49.3} \\
Claude 3.5 Sonnet (new)$^3$ & 71.5 & 48.8 & \gca{60.1} \\
Claude 3.7 Sonnet$^4$ & 78.3 & 41.2 & \gca{59.8} \\
Claude 3.7 Sonnet + optimized prompt $^4$ & 81.2 & 58.4 & \gca{69.8} \\
\bottomrule
\end{tabular}
\vspace{-3mm}
\end{table}

These results across both benchmarks demonstrate that our \pipeline{} approach for generating synthetic multi-turn data through simulated agent-human interplay is highly effective. Models trained on this data consistently outperform open-source baselines and on par with proprietary models, with particularly strong performance in multi-turn scenarios. Importantly, our approach enables smaller models to achieve competitive or superior performance compared to much larger models, highlighting the efficiency and effectiveness of our data generation methodology.

\subsection{Consistency \& Stability Experiments}

\begin{figure}[h]
\centering
\includegraphics[width=0.9\linewidth]{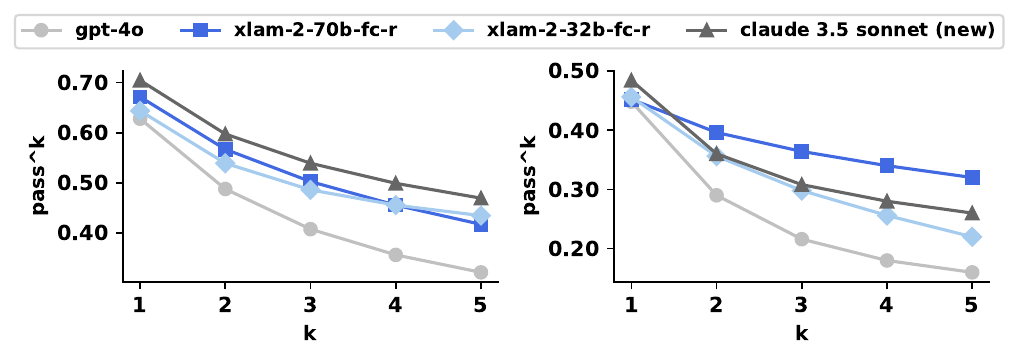}
\caption{\small{Pass\^{}k curves measuring the probability that all 5 independent trials succeed for a given task, averaged across all tasks for $\tau$-retail (left) and $\tau$-airline (right) domains. Higher value indicates consistency of the models.}}
\label{fig:pass_k_curve}
\end{figure}

We plot the \textbf{pass\^{}k} curves \cite{yao2024tau} in \autoref{fig:pass_k_curve} on $\tau$-bench in the default \textit{naive} user LM setting. pass\^{}k is defined as the chance of all $k$ i.i.d. task trials being successful, averaged across all tasks.  As $k$ increases, we see less drop in success rate (SR) for our models. Notably on the more complex airline domain, \modelname{} has higher pass\^{}5 score than Claude, despite having a slightly lower pass\^{}1 suggesting higher reliability and consistency across multiple trials. This is a critical property for deployment in real-world applications, where consistent performance is essential.

Next, we adopt the BoN user LM setting (introduced in \autoref{sec:simulation}) to assess its effectiveness in producing more stable results across trials. Although this enhancement is applied to the user LM, \autoref{tab:stability_results} highlights the improved success rate and reduced variance in models utilizing the BoN user simulation. This suggests that enhancing the user simulation strategy with a simple self-critiquing mechanism can not only increase stability  but also improve agent performance.

\begin{table}[h]
\centering
\tiny
\caption{\small{The Success Rate (SR) measured across 5 trials on the Retail domain of $\tau$-bench using \texttt{gpt-4o} and \modelname~ as the test assistants. The average success rate is higher with lower variance using \textit{BoN} based user simulation, indicative of a more stable evaluation. }}
\renewcommand\arraystretch{1}  %
\begin{tabular}{l|ccccc|c|c}
\toprule
\textbf{Model (User LM setting)} & \textbf{t1} & \textbf{t2} & \textbf{t3} & \textbf{t4} & \textbf{t5} & \textbf{SR Average} & \textbf{SR Variance} \\ 
\midrule
\texttt{gpt-4o} (Naive) & 61.7 & 57.4 & 65.2 & 65.2 & 64.4 & 62.8 &  11.1 \\
\texttt{gpt-4o} (BoN) & 65.2 & 69.6 & 67.0 & 66.1 & 67.0 & \textbf{67.0} & \textbf{2.6} \\
\modelname ~(Naive) & 69.6 & 65.2 & 62.6 & 68.7 & 69.6 & 67.1 & 9.7  \\
\modelname ~(BoN) & 66.9 & 71.3 & 68.7 & 66.9 & 70.4  & \textbf{68.8} & \textbf{4.0} \\
\bottomrule
\end{tabular}
\label{tab:stability_results}
\end{table}

\subsection{In-Depth Analysis of Model Behavior}
\label{sec:model_behavior_analysis}

\begin{wrapfigure}[11]{r}{.45\textwidth}
\vspace{-1.4cm}
\includegraphics[width=.45\textwidth]{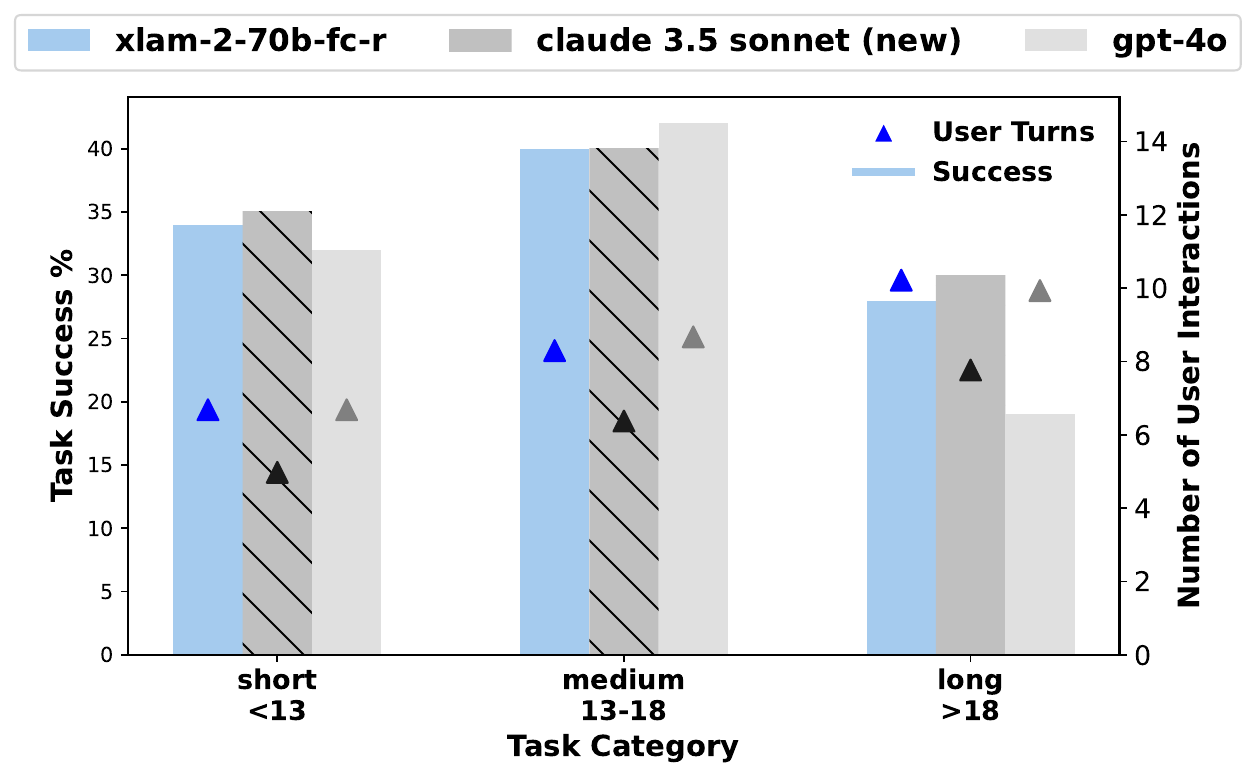}
\caption{\small{Performance/efficiency comparisons of \modelname{} with frontier models on $\tau$-bench.}}
\label{fig:turn_acc}
\end{wrapfigure}

To better understand the behavior of our trained models, we perform an in-depth investigation of the tasks solved by  \modelname{} and a state-of-the-art model Claude 3.5 Sonnet (new) on $\tau$-bench. We categorize tasks into `short', `medium' and `long' based on the number of turns required by Claude 3.5 to solve each task across a union of 8 trials. This categorization is derived by calculating the 33rd and 66th percentiles of the number of turns.  
From \autoref{fig:turn_acc} we observe that particularly on the `long' task category, the success rate for \modelname{} is much higher than \texttt{gpt-4o} but lags behind Claude.
Further, we assess the efficiency of the agent by measuring the number of interactions needed with the simulated user for the agent to fully comprehend the intent and successfully complete the task. The plot reveals that \modelname{} is at par with \texttt{gpt-4o} but requires more interactions compared to Claude, which can be attributed to its method of retrieving user details in stages, necessitating more turns. These observations suggest potential areas for improvement in future iterations.

\section{Discussion}
\label{sec:conclusion}

\textbf{Conclusion. } We introduced \pipeline, a two-phase framework for generating high-quality multi-turn agent data through simulated human-agent interactions. By decoupling the creation of detailed task blueprints from the simulation of conversational trajectories, our approach ensures both structural correctness and natural dialogue dynamics. Experiments on $\tau$-bench and BFCL v3 demonstrate that models trained on our synthetic data outperform existing baselines, with even smaller models showing competitive performance in multi-turn scenarios. Moreover, our stabilization techniques yield more consistent and reliable agent behavior. By open-sourcing our synthetic data and trained models, we aim to foster further advances in AI agent development.

\textbf{Limitations and future directions. } 
Despite its advantages, \pipeline~has limitations that present opportunities for future research. First, while our Best-of-N sampling and self-critique mechanisms reduce human user simulation variance, some stochasticity in human behavior remains; more deterministic simulation methods or refined filtering metrics could further stabilize the process. Second, our current approach discards failed trajectories in the second phase, yet these cases may offer valuable insights; future work could leverage such failures as additional contrastive signal during model training. Third, the multi-stage validation process, though effective, incurs computational overhead; developing more efficient validation or adaptive sampling strategies could improve scalability. Finally, extending our approach to additional domains and incorporating self-improvement through reinforcement learning are promising directions for future work.

\bibliography{main}

\begin{thebibliography}{56}
\providecommand{\natexlab}[1]{#1}
\providecommand{\url}[1]{\texttt{#1}}
\expandafter\ifx\csname urlstyle\endcsname\relax
  \providecommand{\doi}[1]{doi: #1}\else
  \providecommand{\doi}{doi: \begingroup \urlstyle{rm}\Url}\fi

\bibitem[Agashe et~al.(2024)Agashe, Han, Gan, Yang, Li, and Wang]{agashe2024agent_s}
S.~Agashe, J.~Han, S.~Gan, J.~Yang, A.~Li, and X.~E. Wang.
\newblock Agent s: An open agentic framework that uses computers like a human.
\newblock \emph{arXiv preprint arXiv:2410.08164}, 2024.

\bibitem[Anthropic(2024)]{claude3.5}
Anthropic.
\newblock Claude 3.5 sonnet, 2024.
\newblock URL \url{https://www.anthropic.com/news/3-5-models-and-computer-use}.

\bibitem[Anthropic(2025{\natexlab{a}})]{claude3.7}
Anthropic.
\newblock Claude 3.7 sonnet, 2025{\natexlab{a}}.
\newblock URL \url{https://www.anthropic.com/news/claude-3-7-sonnet}.

\bibitem[Anthropic(2025{\natexlab{b}})]{claude_think}
Anthropic.
\newblock Claude think tool, 2025{\natexlab{b}}.
\newblock URL \url{https://www.anthropic.com/engineering/claude-think-tool}.

\bibitem[Antoniades et~al.(2024)Antoniades, {\"O}rwall, Zhang, Xie, Goyal, and Wang]{antoniades2024swe}
A.~Antoniades, A.~{\"O}rwall, K.~Zhang, Y.~Xie, A.~Goyal, and W.~Wang.
\newblock Swe-search: Enhancing software agents with monte carlo tree search and iterative refinement.
\newblock \emph{arXiv preprint arXiv:2410.20285}, 2024.

\bibitem[Arcadinho et~al.(2024)Arcadinho, Apar{\'\i}cio, and Almeida]{arcadinho2024automated}
S.~Arcadinho, D.~Apar{\'\i}cio, and M.~Almeida.
\newblock Automated test generation to evaluate tool-augmented llms as conversational ai agents.
\newblock \emph{arXiv preprint arXiv:2409.15934}, 2024.

\bibitem[Bahdanau et~al.(2024)Bahdanau, Gontier, Huang, Kamalloo, Pardinas, Pich{\'e}, Scholak, Shliazhko, Tremblay, Ghanem, et~al.]{bahdanau2024tapeagents}
D.~Bahdanau, N.~Gontier, G.~Huang, E.~Kamalloo, R.~Pardinas, A.~Pich{\'e}, T.~Scholak, O.~Shliazhko, J.~P. Tremblay, K.~Ghanem, et~al.
\newblock Tapeagents: a holistic framework for agent development and optimization.
\newblock \emph{arXiv preprint arXiv:2412.08445}, 2024.

\bibitem[Bi et~al.(2024)Bi, Han, Liu, Tang, and Wang]{bi2024forest}
Z.~Bi, K.~Han, C.~Liu, Y.~Tang, and Y.~Wang.
\newblock Forest-of-thought: Scaling test-time compute for enhancing llm reasoning.
\newblock \emph{arXiv preprint arXiv:2412.09078}, 2024.

\bibitem[Cai et~al.(2023)Cai, Wang, Ma, Chen, and Zhou]{cai2023large}
T.~Cai, X.~Wang, T.~Ma, X.~Chen, and D.~Zhou.
\newblock Large language models as tool makers.
\newblock \emph{arXiv preprint arXiv:2305.17126}, 2023.

\bibitem[{CAMEL-AI.org}(2025)]{owl2025}
{CAMEL-AI.org}.
\newblock Owl: Optimized workforce learning for general multi-agent assistance in real-world task automation.
\newblock \url{https://github.com/camel-ai/owl}, 2025.
\newblock Accessed: 2025-03-07.

\bibitem[Chen et~al.(2025{\natexlab{a}})Chen, sunhaoze, Li, Yang, Liang, KeerLu, CUI, Zhang, Zhou, and weipeng chen]{chen2025facilitating}
M.~Chen, sunhaoze, T.~Li, F.~Yang, H.~Liang, KeerLu, B.~CUI, W.~Zhang, Z.~Zhou, and weipeng chen.
\newblock Facilitating multi-turn function calling for {LLM}s via compositional instruction tuning.
\newblock In \emph{The Thirteenth International Conference on Learning Representations}, 2025{\natexlab{a}}.
\newblock URL \url{https://openreview.net/forum?id=owP2mymrTD}.

\bibitem[Chen et~al.(2024)Chen, Liu, Wang, Zhang, Liu, Lin, Chen, and Zhao]{chen2024agent}
Z.~Chen, K.~Liu, Q.~Wang, W.~Zhang, J.~Liu, D.~Lin, K.~Chen, and F.~Zhao.
\newblock Agent-flan: Designing data and methods of effective agent tuning for large language models.
\newblock \emph{arXiv preprint arXiv:2403.12881}, 2024.

\bibitem[Chen et~al.(2025{\natexlab{b}})Chen, Li, Huang, Du, Fang, and Zhou]{chen2025atlas}
Z.~Chen, M.~Li, Y.~Huang, Y.~Du, M.~Fang, and T.~Zhou.
\newblock Atlas: Agent tuning via learning critical steps.
\newblock \emph{arXiv preprint arXiv:2503.02197}, 2025{\natexlab{b}}.

\bibitem[Cognition(2025)]{apt}
S.~Cognition.
\newblock Apt-1 blog, 2025.
\newblock URL \url{https://www.scaledcognition.com/blog/apt-1}.

\bibitem[Dao(2023)]{dao2023flashattention2fasterattentionbetter}
T.~Dao.
\newblock Flashattention-2: Faster attention with better parallelism and work partitioning, 2023.
\newblock URL \url{https://arxiv.org/abs/2307.08691}.

\bibitem[Dubey et~al.(2024)Dubey, Jauhri, Pandey, Kadian, Al-Dahle, Letman, Mathur, Schelten, Yang, Fan, et~al.]{dubey2024llama}
A.~Dubey, A.~Jauhri, A.~Pandey, A.~Kadian, A.~Al-Dahle, A.~Letman, A.~Mathur, A.~Schelten, A.~Yang, A.~Fan, et~al.
\newblock The llama 3 herd of models.
\newblock \emph{arXiv preprint arXiv:2407.21783}, 2024.

\bibitem[Ge et~al.(2024)Ge, Chan, Wang, Yu, Mi, and Yu]{ge2024scalingsyntheticdatacreation}
T.~Ge, X.~Chan, X.~Wang, D.~Yu, H.~Mi, and D.~Yu.
\newblock Scaling synthetic data creation with 1,000,000,000 personas, 2024.
\newblock URL \url{https://arxiv.org/abs/2406.20094}.

\bibitem[Hayati et~al.(2024)Hayati, Jung, Bodding-Long, Kar, Sethy, Kim, and Kang]{hayati2024chain}
S.~A. Hayati, T.~Jung, T.~Bodding-Long, S.~Kar, A.~Sethy, J.-K. Kim, and D.~Kang.
\newblock Chain-of-instructions: Compositional instruction tuning on large language models.
\newblock \emph{arXiv preprint arXiv:2402.11532}, 2024.

\bibitem[Huang et~al.(2025)Huang, Prabhakar, Dhawan, Mao, Wang, Savarese, Xiong, Laban, and Wu]{huang2025crmarenaunderstandingcapacityllm}
K.-H. Huang, A.~Prabhakar, S.~Dhawan, Y.~Mao, H.~Wang, S.~Savarese, C.~Xiong, P.~Laban, and C.-S. Wu.
\newblock Crmarena: Understanding the capacity of llm agents to perform professional crm tasks in realistic environments, 2025.
\newblock URL \url{https://arxiv.org/abs/2411.02305}.

\bibitem[Levi and Kadar(2025)]{levi2025intellagent}
E.~Levi and I.~Kadar.
\newblock Intellagent: A multi-agent framework for evaluating conversational ai systems.
\newblock \emph{arXiv preprint arXiv:2501.11067}, 2025.

\bibitem[Li et~al.(2023)Li, Hammoud, Itani, Khizbullin, and Ghanem]{li2023camel}
G.~Li, H.~A. A.~K. Hammoud, H.~Itani, D.~Khizbullin, and B.~Ghanem.
\newblock Camel: Communicative agents for "mind" exploration of large language model society.
\newblock In \emph{Thirty-seventh Conference on Neural Information Processing Systems}, 2023.

\bibitem[Li et~al.(2025)Li, Zou, and Liu]{li2025torlscalingtoolintegratedrl}
X.~Li, H.~Zou, and P.~Liu.
\newblock Torl: Scaling tool-integrated rl, 2025.
\newblock URL \url{https://arxiv.org/abs/2503.23383}.

\bibitem[Li et~al.(2024)Li, Li, Wang, Jiang, Zhang, Zheng, Wang, Zheng, Xie, Yu, et~al.]{li2024benchmarking}
Y.~Li, Y.~Li, X.~Wang, Y.~Jiang, Z.~Zhang, X.~Zheng, H.~Wang, H.-T. Zheng, P.~Xie, P.~S. Yu, et~al.
\newblock Benchmarking multimodal retrieval augmented generation with dynamic vqa dataset and self-adaptive planning agent.
\newblock \emph{arXiv preprint arXiv:2411.02937}, 2024.

\bibitem[Liu et~al.(2024{\natexlab{a}})Liu, Huang, Zeng, Hao, Yu, Li, Wang, Gan, Liu, Yu, et~al.]{liu2024toolace}
W.~Liu, X.~Huang, X.~Zeng, X.~Hao, S.~Yu, D.~Li, S.~Wang, W.~Gan, Z.~Liu, Y.~Yu, et~al.
\newblock Toolace: Winning the points of llm function calling.
\newblock \emph{arXiv preprint arXiv:2409.00920}, 2024{\natexlab{a}}.

\bibitem[Liu et~al.(2023)Liu, Zhang, Asadi, Liu, Zhao, Sabach, and Fakoor]{liu2023tail}
Z.~Liu, J.~Zhang, K.~Asadi, Y.~Liu, D.~Zhao, S.~Sabach, and R.~Fakoor.
\newblock Tail: Task-specific adapters for imitation learning with large pretrained models.
\newblock \emph{arXiv preprint arXiv:2310.05905}, 2023.

\bibitem[Liu et~al.(2024{\natexlab{b}})Liu, Hoang, Zhang, Zhu, Lan, Tan, Yao, Liu, Feng, RN, et~al.]{liu2024apigen}
Z.~Liu, T.~Hoang, J.~Zhang, M.~Zhu, T.~Lan, J.~Tan, W.~Yao, Z.~Liu, Y.~Feng, R.~RN, et~al.
\newblock Apigen: Automated pipeline for generating verifiable and diverse function-calling datasets.
\newblock \emph{Advances in Neural Information Processing Systems}, 37:\penalty0 54463--54482, 2024{\natexlab{b}}.

\bibitem[Loshchilov and Hutter(2017)]{loshchilov2017decoupled}
I.~Loshchilov and F.~Hutter.
\newblock Decoupled weight decay regularization.
\newblock \emph{arXiv preprint arXiv:1711.05101}, 2017.

\bibitem[Lu et~al.(2024)Lu, Holleis, Zhang, Aumayer, Nan, Bai, Ma, Ma, Li, Yin, et~al.]{lu2024toolsandbox}
J.~Lu, T.~Holleis, Y.~Zhang, B.~Aumayer, F.~Nan, F.~Bai, S.~Ma, S.~Ma, M.~Li, G.~Yin, et~al.
\newblock Toolsandbox: A stateful, conversational, interactive evaluation benchmark for llm tool use capabilities.
\newblock \emph{arXiv preprint arXiv:2408.04682}, 2024.

\bibitem[Mitra et~al.(2024)Mitra, Patel, Chakrabarty, and Baral]{mitra2024agentinstruct}
A.~Mitra, S.~Patel, T.~Chakrabarty, and C.~Baral.
\newblock Agentinstruct: An agentic framework for generating high-quality synthetic instruction data.
\newblock \emph{arXiv preprint arXiv:2402.12360}, 2024.

\bibitem[Pan et~al.(2024)Pan, Wang, Neubig, Jaitly, Ji, Suhr, and Zhang]{pan2024training}
J.~Pan, X.~Wang, G.~Neubig, N.~Jaitly, H.~Ji, A.~Suhr, and Y.~Zhang.
\newblock Training software engineering agents and verifiers with swe-gym.
\newblock \emph{arXiv preprint arXiv:2412.21139}, 2024.

\bibitem[Pan et~al.(2025)Pan, Shar, Pfau, Talwalkar, He, and Chen]{pan2025benchmarkstalkreevaluatingcode}
J.~Pan, R.~Shar, J.~Pfau, A.~Talwalkar, H.~He, and V.~Chen.
\newblock When benchmarks talk: Re-evaluating code llms with interactive feedback, 2025.
\newblock URL \url{https://arxiv.org/abs/2502.18413}.

\bibitem[Park et~al.(2023)Park, O'Brien, Cai, Morris, Liang, and Bernstein]{park2023generative}
J.~S. Park, J.~O'Brien, C.~J. Cai, M.~R. Morris, P.~Liang, and M.~S. Bernstein.
\newblock Generative agents: Interactive simulacra of human behavior.
\newblock In \emph{Proceedings of the 36th annual acm symposium on user interface software and technology}, pages 1--22, 2023.

\bibitem[Qin et~al.(2024)Qin, Hu, Lin, Chen, Ding, Cui, Zeng, Zhou, Huang, Xiao, et~al.]{qin2024tool}
Y.~Qin, S.~Hu, Y.~Lin, W.~Chen, N.~Ding, G.~Cui, Z.~Zeng, X.~Zhou, Y.~Huang, C.~Xiao, et~al.
\newblock Tool learning with foundation models.
\newblock \emph{ACM Computing Surveys}, 57\penalty0 (4):\penalty0 1--40, 2024.

\bibitem[Qwen et~al.(2025)Qwen, :, Yang, Yang, Zhang, Hui, Zheng, Yu, Li, Liu, Huang, Wei, Lin, Yang, Tu, Zhang, Yang, Yang, Zhou, Lin, Dang, Lu, Bao, Yang, Yu, Li, Xue, Zhang, Zhu, Men, Lin, Li, Tang, Xia, Ren, Ren, Fan, Su, Zhang, Wan, Liu, Cui, Zhang, and Qiu]{qwen2025qwen25technicalreport}
Qwen, :, A.~Yang, B.~Yang, B.~Zhang, B.~Hui, B.~Zheng, B.~Yu, C.~Li, D.~Liu, F.~Huang, H.~Wei, H.~Lin, J.~Yang, J.~Tu, J.~Zhang, J.~Yang, J.~Yang, J.~Zhou, J.~Lin, K.~Dang, K.~Lu, K.~Bao, K.~Yang, L.~Yu, M.~Li, M.~Xue, P.~Zhang, Q.~Zhu, R.~Men, R.~Lin, T.~Li, T.~Tang, T.~Xia, X.~Ren, X.~Ren, Y.~Fan, Y.~Su, Y.~Zhang, Y.~Wan, Y.~Liu, Z.~Cui, Z.~Zhang, and Z.~Qiu.
\newblock Qwen2.5 technical report, 2025.
\newblock URL \url{https://arxiv.org/abs/2412.15115}.

\bibitem[Rasley et~al.(2020)Rasley, Rajbhandari, Ruwase, and He]{10.1145/3394486.3406703}
J.~Rasley, S.~Rajbhandari, O.~Ruwase, and Y.~He.
\newblock Deepspeed: System optimizations enable training deep learning models with over 100 billion parameters.
\newblock In \emph{Proceedings of the 26th ACM SIGKDD International Conference on Knowledge Discovery \& Data Mining}, KDD '20, page 3505–3506, New York, NY, USA, 2020. Association for Computing Machinery.
\newblock ISBN 9781450379984.
\newblock \doi{10.1145/3394486.3406703}.
\newblock URL \url{https://doi.org/10.1145/3394486.3406703}.

\bibitem[Schick et~al.(2023)Schick, Dwivedi-Yu, Dess{\`\i}, Raileanu, Lomeli, Hambro, Zettlemoyer, Cancedda, and Scialom]{schick2023toolformer}
T.~Schick, J.~Dwivedi-Yu, R.~Dess{\`\i}, R.~Raileanu, M.~Lomeli, E.~Hambro, L.~Zettlemoyer, N.~Cancedda, and T.~Scialom.
\newblock Toolformer: Language models can teach themselves to use tools.
\newblock \emph{Advances in Neural Information Processing Systems}, 36:\penalty0 68539--68551, 2023.

\bibitem[Sengupta et~al.(2024)Sengupta, Curtis, Mallipeddi, Mathur, Ross, and Gou]{sengupta2024mag}
S.~Sengupta, K.~Curtis, A.~Mallipeddi, A.~Mathur, J.~Ross, and L.~Gou.
\newblock Mag-v: A multi-agent framework for synthetic data generation and verification.
\newblock \emph{arXiv preprint arXiv:2412.04494}, 2024.

\bibitem[Shim et~al.(2025)Shim, Seo, Lim, and Jo]{shim2025tooldial}
J.~Shim, G.~Seo, C.~Lim, and Y.~Jo.
\newblock Tooldial: Multi-turn dialogue generation method for tool-augmented language models.
\newblock In \emph{The Thirteenth International Conference on Learning Representations}, 2025.
\newblock URL \url{https://openreview.net/forum?id=J1J5eGJsKZ}.

\bibitem[Shinn et~al.(2023)Shinn, Cassano, Gopinath, Narasimhan, and Yao]{shinn2023reflexion}
N.~Shinn, F.~Cassano, A.~Gopinath, K.~R. Narasimhan, and S.~Yao.
\newblock Reflexion: language agents with verbal reinforcement learning.
\newblock In \emph{Thirty-seventh Conference on Neural Information Processing Systems}, 2023.
\newblock URL \url{https://openreview.net/forum?id=vAElhFcKW6}.

\bibitem[Sirdeshmukh et~al.(2025)Sirdeshmukh, Deshpande, Mols, Jin, Cardona, Lee, Kritz, Primack, Yue, and Xing]{sirdeshmukh2025multichallenge}
V.~Sirdeshmukh, K.~Deshpande, J.~Mols, L.~Jin, E.-Y. Cardona, D.~Lee, J.~Kritz, W.~Primack, S.~Yue, and C.~Xing.
\newblock Multichallenge: A realistic multi-turn conversation evaluation benchmark challenging to frontier llms.
\newblock \emph{arXiv preprint arXiv:2501.17399}, 2025.

\bibitem[Su et~al.(2025)Su, Sun, Yoon, Yin, Yu, and Ar{\i}k]{su2025learn}
H.~Su, R.~Sun, J.~Yoon, P.~Yin, T.~Yu, and S.~{\"O}. Ar{\i}k.
\newblock Learn-by-interact: A data-centric framework for self-adaptive agents in realistic environments.
\newblock \emph{arXiv preprint arXiv:2501.10893}, 2025.

\bibitem[Tang et~al.(2024)Tang, Pang, Liu, Tang, Ye, Dong, Wang, and Chen]{tang2024synthesizing}
S.~Tang, X.~Pang, Z.~Liu, B.~Tang, R.~Ye, X.~Dong, Y.~Wang, and S.~Chen.
\newblock Synthesizing post-training data for llms through multi-agent simulation.
\newblock \emph{arXiv preprint arXiv:2410.14251}, 2024.

\bibitem[Wang et~al.(2024)Wang, Zhou, Wen, Mo, Zhang, Lin, Jin, Wang, Zhang, and Peng]{wang2024hammerbench}
J.~Wang, J.~Zhou, M.~Wen, X.~Mo, H.~Zhang, Q.~Lin, C.~Jin, X.~Wang, W.~Zhang, and Q.~Peng.
\newblock Hammerbench: Fine-grained function-calling evaluation in real mobile device scenarios.
\newblock \emph{arXiv preprint arXiv:2412.16516}, 2024.

\bibitem[W{\"o}lflein et~al.(2025)W{\"o}lflein, Ferber, Truhn, Arandjelovi{\'c}, and Kather]{wölflein2025llm}
G.~W{\"o}lflein, D.~Ferber, D.~Truhn, O.~Arandjelovi{\'c}, and J.~N. Kather.
\newblock Llm agents making agent tools.
\newblock \emph{arXiv preprint arXiv:2502.11705}, 2025.

\bibitem[Yan et~al.(2024)Yan, Mao, Ji, Zhang, Patil, Stoica, and Gonzalez]{berkeley-function-calling-leaderboard}
F.~Yan, H.~Mao, C.~C.-J. Ji, T.~Zhang, S.~G. Patil, I.~Stoica, and J.~E. Gonzalez.
\newblock Berkeley function calling leaderboard.
\newblock 2024.

\bibitem[Yang et~al.(2023)Yang, Prabhakar, Yao, Pei, and Narasimhan]{yang2023language}
J.~Yang, A.~Prabhakar, S.~Yao, K.~Pei, and K.~R. Narasimhan.
\newblock Language agents as hackers: Evaluating cybersecurity skills with capture the flag.
\newblock In \emph{Multi-Agent Security Workshop @ NeurIPS'23}, 2023.
\newblock URL \url{https://openreview.net/forum?id=KOZwk7BFc3}.

\bibitem[Yang et~al.(2024)Yang, Prabhakar, Narasimhan, and Yao]{yang2024intercode}
J.~Yang, A.~Prabhakar, K.~Narasimhan, and S.~Yao.
\newblock Intercode: Standardizing and benchmarking interactive coding with execution feedback.
\newblock \emph{Advances in Neural Information Processing Systems}, 36, 2024.

\bibitem[Yang et~al.(2025)Yang, Ye, Li, Yuan, Zhang, Tu, Li, and Yang]{yang2025lighthouse}
R.~Yang, F.~Ye, J.~Li, S.~Yuan, Y.~Zhang, Z.~Tu, X.~Li, and D.~Yang.
\newblock The lighthouse of language: Enhancing llm agents via critique-guided improvement.
\newblock \emph{arXiv preprint arXiv:2503.16024}, 2025.

\bibitem[Yao et~al.(2024)Yao, Shinn, Razavi, and Narasimhan]{yao2024tau}
S.~Yao, N.~Shinn, P.~Razavi, and K.~Narasimhan.
\newblock Tau-bench: A benchmark for tool-agent-user interaction in real-world domains.
\newblock \emph{arXiv preprint arXiv:2406.12045}, 2024.

\bibitem[Yin et~al.(2025)Yin, Wang, Hsu, Yan, Jiang, Chen, Gu, Le, Chang, Lee, Palangi, and Pfister]{yin2025magnetmultiturntoolusedata}
F.~Yin, Z.~Wang, I.-H. Hsu, J.~Yan, K.~Jiang, Y.~Chen, J.~Gu, L.~T. Le, K.-W. Chang, C.-Y. Lee, H.~Palangi, and T.~Pfister.
\newblock Magnet: Multi-turn tool-use data synthesis and distillation via graph translation, 2025.
\newblock URL \url{https://arxiv.org/abs/2503.07826}.

\bibitem[Zeng et~al.(2025)Zeng, Ding, Wang, Liu, Ning, Hou, Huang, Qin, and Liu]{zeng2025boosting}
Y.~Zeng, X.~Ding, Y.~Wang, W.~Liu, W.~Ning, Y.~Hou, X.~Huang, B.~Qin, and T.~Liu.
\newblock Boosting tool use of large language models via iterative reinforced fine-tuning.
\newblock \emph{arXiv preprint arXiv:2501.09766}, 2025.

\bibitem[Zhang et~al.(2024{\natexlab{a}})Zhang, Lan, Zhu, Liu, Hoang, Kokane, Yao, Tan, Prabhakar, Chen, et~al.]{zhang2024xlam}
J.~Zhang, T.~Lan, M.~Zhu, Z.~Liu, T.~Hoang, S.~Kokane, W.~Yao, J.~Tan, A.~Prabhakar, H.~Chen, et~al.
\newblock xlam: A family of large action models to empower ai agent systems.
\newblock \emph{arXiv preprint arXiv:2409.03215}, 2024{\natexlab{a}}.

\bibitem[Zhang et~al.(2025)Zhang, Hoang, Zhu, Liu, Wang, Awalgaonkar, Prabhakar, Chen, Yao, Liu, et~al.]{zhang2025actionstudio}
J.~Zhang, T.~Hoang, M.~Zhu, Z.~Liu, S.~Wang, T.~Awalgaonkar, A.~Prabhakar, H.~Chen, W.~Yao, Z.~Liu, et~al.
\newblock Actionstudio: A lightweight framework for data and training of action models.
\newblock \emph{arXiv preprint arXiv:2503.22673}, 2025.

\bibitem[Zhang et~al.(2024{\natexlab{b}})Zhang, Yao, Liu, Feng, Liu, Rithesh, Lan, Li, Lou, Xu, et~al.]{zhang2024diversity}
K.~Zhang, W.~Yao, Z.~Liu, Y.~Feng, Z.~Liu, R.~Rithesh, T.~Lan, L.~Li, R.~Lou, J.~Xu, et~al.
\newblock Diversity empowers intelligence: Integrating expertise of software engineering agents.
\newblock In \emph{The Thirteenth International Conference on Learning Representations}, 2024{\natexlab{b}}.

\bibitem[Zhang et~al.(2024{\natexlab{c}})Zhang, Lu, and Jaitly]{zhang-etal-2024-probing}
Y.~Zhang, J.~Lu, and N.~Jaitly.
\newblock Probing the multi-turn planning capabilities of {LLM}s via 20 question games.
\newblock In L.-W. Ku, A.~Martins, and V.~Srikumar, editors, \emph{Proceedings of the 62nd Annual Meeting of the Association for Computational Linguistics (Volume 1: Long Papers)}, pages 1495--1516, Bangkok, Thailand, Aug. 2024{\natexlab{c}}. Association for Computational Linguistics.
\newblock \doi{10.18653/v1/2024.acl-long.82}.
\newblock URL \url{https://aclanthology.org/2024.acl-long.82/}.

\bibitem[Zheng et~al.(2024)Zheng, Zhang, Zhang, Ye, Luo, Feng, and Ma]{zheng2024llamafactory}
Y.~Zheng, R.~Zhang, J.~Zhang, Y.~Ye, Z.~Luo, Z.~Feng, and Y.~Ma.
\newblock Llamafactory: Unified efficient fine-tuning of 100+ language models.
\newblock In \emph{Proceedings of the 62nd Annual Meeting of the Association for Computational Linguistics (Volume 3: System Demonstrations)}, Bangkok, Thailand, 2024. Association for Computational Linguistics.
\newblock URL \url{http://arxiv.org/abs/2403.13372}.

\end{thebibliography}
\bibliographystyle{abbrvnat}

\newpage
\appendix
\section{Benchmarks Description}
\label{appx:benchmarks}
\begin{itemize}[leftmargin=*]
    \item \textbf{BFCL v3}: It introduces comprehensive evaluation across single-turn, multi-turn, and multi-step function calling scenarios. BFCL v3 evaluates models on their ability to understand user requests, select appropriate functions, generate valid parameters, and interpret function outputs across multiple interaction turns. The benchmark uses a weighted average of different evaluation categories to provide an overall accuracy score.
    \item \textbf{$\boldsymbol{\tau}$-bench}: It measures an agent's ability to interact with simulated human users (powered by language models) and programmatic APIs while following domain-specific policies. $\tau$-bench emulates dynamic conversations across multiple domains, including retail and airline customer service, requiring agents to maintain context across turns, understand user intents, and follow complex domain-specific rules. The benchmark emphasizes the importance of multi-turn interactions and policy adherence in real-world applications.
\end{itemize}

\section{Prompts}
\label{appx:prompts}

The prompts used across the various stages of \pipeline{} implemented for $\tau$-bench are shown here -- Task Configuration Generation (\autoref{fig:generation_prompt}), Alignment Validation (\autoref{fig:validation_prompt}), Final Semantic Review (\autoref{fig:review_prompt}), Trajectory Collection (\autoref{fig:traj_prompt}), Stabilized Human Simulation (\autoref{fig:bon_prompt}).

\begin{figure}[h]
\centering
\begin{tcolorbox}[
    width=\textwidth,
    left=0.5mm,
    right=0.5mm,
    top=0.5mm,
    bottom=0.5mm,
    title=Task Configuration Generation Prompt,
    center title,
    fonttitle=\tiny\ttfamily]
\tiny
\#\# Instructions\\
Generate a task instruction that mimics realistic human users and their intentions, such as with different personality and goals. The task instruction should be followed by `actions' which is a list of the tool\_calls to be taken to solve this task and `outputs' which is a list of the answers to specific information requests made by the user. Think step by step to come up with the action(s) and the corresponding tool\_call(s) translating this thought that would be necessary to fulfill the user's request or solve their intentions. Focus on common retail scenarios following the provided task instruction guidelines.\\

\#\# Guidelines for Generating Task Instruction ($q$)\\
\{task\_rules + domain\_rules\}\\

\#\#\# User Data\\
\{sampled\_user\_details\}\\

\#\#\# Order Data\\
\{sampled\_orders\}\\

\#\# Guidelines for generating Groundtruth Actions ($a_{gt}$)
\begin{enumerate}[itemsep=1pt,topsep=0pt,leftmargin=1.5em]
    \item The main focus is to generate actions that can modify the underlying database.
    \item For actions that do not modify the database like specific information requests, scan the provided User Data directly and append only the answer in `outputs' ($o_{gt}$). Do not make separate tool calls for this in `actions'.
    \item Include multiple tool calls when the scenario requires multiple steps or modifications.
    \item Provide precise tool calls with all necessary parameters for each action.
    \item Ensure all actions adhere to retail policies and common sense practices.\\
\end{enumerate}

\#\# Tools\\
The available tool combination in Python format is as follows:\\
\{sampled\_tools\}\\

\#\# Output Format\\
Generate your response according to the following format. Enclose the thought process within `<thought></thought>' tags, and the final structured response within `<answer></answer>' tags. The structured response should be in strict JSON format, without any additional comments or explanations.\\

\#\# Example Tasks\\
\{example\}\\

Do not directly copy instruction and the action patterns from the examples. Ground the generation from the above provided data.

Generate the task now.
\end{tcolorbox}
\caption{Task configuration generation prompt for retail domain of $\tau$-bench.}
\label{fig:generation_prompt}
\end{figure}

\begin{figure}[h]
\centering
\begin{tcolorbox}[
    width=\textwidth,
    left=0.5mm,
    right=0.5mm,
    top=0.5mm,
    bottom=0.5mm,
    title=Task Alignment Validation Prompt,
    center title,
    fonttitle=\tiny\ttfamily]
\tiny
You are an AI judge and your goal is to judge the quality and validity of the provided task object based on the guidelines, following the rubric.\\

\#\# Guidelines
\begin{itemize}[itemsep=1pt,topsep=0pt,leftmargin=2.5em]
\item The task object contains an `intent' ($q$) from a user, `actions' ($a_{gt}$), and `outputs' ($o_{gt}$).
\item The `actions' correspond to the tool\_calls made by an AI assistant to satisfy the instruction.
\item A description of the `tools' available to the AI assistant is provided. 
\item The `diff\_patch' is the difference in the database state after the tool\_calls are made. It should only reflect changes corresponding to the `intent'. There should be no extraneous changes. If the `diff\_patch' is empty, it means that the tool\_calls did not change the database state, which is possible if the instruction was to provide information only.
\item Perform a brief reflection on the task based on the below Rubrics.
\item Think step-by-step to generate a score of 0 or 1 for each of these criteria (1 means follows criterion and 0 means does not)\\
\end{itemize}

\#\# Rubric
\begin{itemize}[itemsep=1pt,topsep=0pt,leftmargin=2.5em]
    \item \textit{Correctness}: Do the actions ($a_{gt}$) accurately implement the instruction ($q$)?
    \item \textit{Completeness}: Is the instruction ($q$) sufficiently detailed, and is it fully addressed by the actions? (Includes rule-based checks).
    \item \textit{Satisfaction}: Do the expected outputs ($o_{gt}$) fulfill any explicit or implicit information requests within the instruction ($q$)?
    \item \textit{Creativity}: Does the task represent a non-trivial, plausible, and potentially interesting scenario within the domain?\\
\end{itemize}

\#\# Task Object\\
\{task\}
\\

\#\# Tools in Python format\\
\{tools\}\\

\#\# Diff Patch\\
\{diff\_patch\}\\

\#\# Output format\\
<scores>\\
\{\{\\
\hspace*{2em}    "reflection": str, <a brief high-level review of the task>\\
\hspace*{2em}    "correctness": int, <0/1>,\\
\hspace*{2em}    "completeness": int, <0/1>,\\
\hspace*{2em}    "satisfaction": int, <0/1>,\\
\hspace*{2em}    "creativity": int, <0/1>,\\
\hspace*{2em}    "total": int, <total score out of 4>\\
\hspace*{2em}    "correction": str, <brief explanation and suggested correction (if needed)>\\
\}\}\\
</scores>
\end{tcolorbox}
\caption{Task alignment validation prompt for $\tau$-bench. This is sent to each LM in the review committee to get their scores, following which we employ majority voting.}
\label{fig:validation_prompt}
\end{figure}

\begin{figure}[h]
\centering
\begin{tcolorbox}[
    width=\textwidth,
    left=0.5mm,
    right=0.5mm,
    top=0.5mm,
    bottom=0.5mm,
    title=Final Semantic Review Prompt,
    center title,
    fonttitle=\tiny\ttfamily]
\tiny
You are responsible for analyzing and summarizing feedback from multiple AI judges. Your primary goal is to provide clear, actionable feedback that will help the generator LLM improve its future outputs. You do not evaluate the task directly; instead, you review and grounding the existing feedback from the AI judges.\\

\#\# Review Process
\begin{itemize}[itemsep=1pt,topsep=0pt,leftmargin=2.5em]
    \item Begin by analyzing individual reflections and scores from each judge.
    \item Summarize common points of agreement or disagreement.
    \item Offer a concise summary of actionable feedback to be sent back to the data generator, which aims to improve the next round of data quality.\\
\end{itemize}

\#\#\# Diff Patch\\
\{diff\_patch\}\\

\#\#\# Generated Task Data\\
\{task\}\\

\#\#\# AI Judges' Feedback\\
\{reviews\}\\

\#\# Output Format\\
Generate your response according to the following format. Enclose the thought process within `<thought></thought>' tags, and the final summary of actionable feedback within `<summary></summary>' tags.
\end{tcolorbox}
\caption{Final semantic review prompt for $\tau$-bench.}
\label{fig:review_prompt}
\end{figure}

\begin{figure}[h]
\centering
\begin{tcolorbox}[
    width=\textwidth,
    left=0.5mm,
    right=0.5mm,
    top=0.5mm,
    bottom=0.5mm,
    title=Trajectory Collection Prompt,
    center title,
    fonttitle=\tiny\ttfamily]
\tiny
You are a detail-oriented user interacting with an AI agent.\\

\#\# Intent\\
\{intent\}\\

\#\# Rules
\begin{itemize}[itemsep=1pt,topsep=0pt,leftmargin=2.5em]
    \item  Generate one line at a time to simulate the user's message.
    \item  Do not give away all the intent at once. Only provide the information that is necessary for the current step.
    \item  Do not hallucinate information that is not provided in the intent. 
    \item  If the intent goal is satisfied, generate `\#\#\#STOP\#\#\#' to end the conversation.
    \item  Do not repeat the exact intent in the conversation. Instead, use your own words to convey the same information.
    \item  Try to make the conversation as natural as possible and stick to the personalities in the intent.
\end{itemize}
\end{tcolorbox}
\caption{Trajectory collection prompt for $\tau$-bench.}
\label{fig:traj_prompt}
\end{figure}

\begin{figure}[h]
\centering
\begin{tcolorbox}[
    width=\textwidth,
    left=0.5mm,
    right=0.5mm,
    top=0.5mm,
    bottom=0.5mm,
    title=BoN User LM Setting Prompt,
    center title,
    fonttitle=\tiny\ttfamily]
\tiny
You are a fair judge and an expert in following details.\\
\\
A human is interacting with a retail assistant to get help on solving their task. You are provided with the description of the human and the task the human wants to accomplish (wrapped with <description></description>), and a candidate response (wrapped with <response></response>) the human wants to give the assistant. Please help the human evaluate this candidate response, give an integer score (ranging from 0 to 10) to indicate the correctness of the response, higher score means better quality.\\

1. If the response includes specific item / order / personal details, and they correctly match the task description you should give full score of 10. If there is some change in details, give a corresponding lower score (more incorrect details gets lower score).\\
2. The response can include any normal conversation otherwise (e.g. asking details, saying \#\#\#STOP\#\#\#) etc. which are all correct responses.\\
3. Additionally, if the candidate\_response keeps the conversation flowing by describing the task clearly / gives information properly then give a high score and if not (e.g. "I don't remember" or unhelpful response) should get a corresponding lower score.\\

<description>
\{description\}
</description>\\

<response>
\{response\}
</response>\\

After scoring using the mentioned guideline, tell me your score, wrap it in <score></score> tags.
\end{tcolorbox}
\caption{Best-of-N (BoN) User LM setting prompt used in the retail domain of $\tau$-bench.}
\label{fig:bon_prompt}
\end{figure}

\end{document}